\pdfoutput=1 %
\documentclass[11pt]{article}
\usepackage[font=libertinus, citestyle=numeric]{kurbanlab}

\DeclareAffiliation{hbku}{%
  College of Science and Engineering, Hamad Bin Khalifa University, Doha, Qatar}

\DeclareAffiliation{tamu}{%
  Department of Computer and Electrical Engineering,
  Texas A\&M University, College Station, TX, USA}

\DeclareAffiliation{iub}{%
  Luddy School of Informatics, Computing, and Engineering,
  Indiana University, Bloomington, IN, USA}

\DeclareAffiliation{uis}{%
  Department of Computer Science, University of Illinois Springfield, Springfield, IL, USA}

\ExplSyntaxOn
\RenewDocumentCommand \KILprintauthors {}
  { \seq_use:Nnnn \g_kil_author_seq { ~and~ } { ,~ } { ,~and~ } }
\ExplSyntaxOff

\usepackage{pifont}
\usepackage{xspace}
\algrenewcommand\algorithmicrequire{\textbf{Input:}}
\algrenewcommand\algorithmicensure{\textbf{Output:}}
\algrenewcommand\algorithmiccomment[1]{\hfill{\color{gray!70!black}$\triangleright$ #1}}
\Crefname{section}{Section}{Sections}
\Crefname{figure}{Figure}{Figures}
\Crefname{table}{Table}{Tables}
\Crefname{equation}{Equation}{Equations}
\Crefname{algorithm}{Algorithm}{Algorithms}

\definecolor{pendred}{RGB}{196,30,30}

\newcommand{\GHURL}{https://github.com/KurbanIntelligenceLab/CivicBench}
\newcommand{\HFURL}{https://huggingface.co/datasets/kurbanintelligencelab/CivicBench}
\newcommand{\PYURL}{https://pypi.org/project/civicbench/}
\usetikzlibrary{svg.path}
\definecolor{hf3A3B45}{HTML}{3A3B45}
\definecolor{hfFF323D}{HTML}{FF323D}
\definecolor{hfFF9D0B}{HTML}{FF9D0B}
\definecolor{hfFFD21E}{HTML}{FFD21E}
\newcommand{\ghlogo}[2]{{\tikz[x=1pt,y=1pt,baseline={([yshift=-0.5ex]current bounding box.center)}]{\begin{scope}[scale=#1/24,yscale=-1]\fill[#2] svg {M 12.5 .75 C 6.146 .75 1 5.896 1 12.25 c 0 5.089 3.292 9.387 7.863 10.91 .575 .101 .79 -.244 .79 -.546 0 -.273 -.014 -1.178 -.014 -2.142 -2.889 .532 -3.636 -.704 -3.866 -1.35 -.13 -.331 -.69 -1.352 -1.18 -1.625 -.402 -.216 -.977 -.748 -.014 -.762 .906 -.014 1.553 .834 1.769 1.179 1.035 1.74 2.688 1.25 3.349 .948 .1 -.747 .402 -1.25 .733 -1.538 -2.559 -.287 -5.232 -1.279 -5.232 -5.678 0 -1.25 .445 -2.285 1.178 -3.09 -.115 -.288 -.517 -1.467 .115 -3.048 0 0 .963 -.302 3.163 1.179 .92 -.259 1.897 -.388 2.875 -.388 .977 0 1.955 .13 2.875 .388 2.2 -1.495 3.162 -1.179 3.162 -1.179 .633 1.581 .23 2.76 .115 3.048 .733 .805 1.179 1.825 1.179 3.09 0 4.413 -2.688 5.39 -5.247 5.678 .417 .36 .776 1.05 .776 2.128 0 1.538 -.014 2.774 -.014 3.162 0 .302 .216 .662 .79 .547 C 20.709 21.637 24 17.324 24 12.25 24 5.896 18.854 .75 12.5 .75 Z};\end{scope}}}}
\newcommand{\hflogo}[1]{{\tikz[x=1pt,y=1pt,baseline={([yshift=-0.5ex]current bounding box.center)}]{\begin{scope}[scale=#1/88,yscale=-1]\fill[hfFFD21E] svg {M 47.21 76.5 a 34.75 34.75 0 1 0 0 -69.5 34.75 34.75 0 0 0 0 69.5 Z};\fill[hfFF9D0B] svg {M 81.96 41.75 a 34.75 34.75 0 1 0 -69.5 0 34.75 34.75 0 0 0 69.5 0 Z m -73.5 0 a 38.75 38.75 0 1 1 77.5 0 38.75 38.75 0 0 1 -77.5 0 Z};\fill[hf3A3B45] svg {M 58.5 32.3 c 1.28 .44 1.78 3.06 3.07 2.38 a 5 5 0 1 0 -6.76 -2.07 c .61 1.15 2.55 -.72 3.7 -.32 Z M 34.95 32.3 c -1.28 .44 -1.79 3.06 -3.07 2.38 a 5 5 0 1 1 6.76 -2.07 c -.61 1.15 -2.56 -.72 -3.7 -.32 Z};\fill[hfFF323D] svg {M 46.96 56.29 c 9.83 0 13 -8.76 13 -13.26 0 -2.34 -1.57 -1.6 -4.09 -.36 -2.33 1.15 -5.46 2.74 -8.9 2.74 -7.19 0 -13 -6.88 -13 -2.38 s 3.16 13.26 13 13.26 Z};\fill[hf3A3B45,even odd rule] svg {M 39.43 54 a 8.7 8.7 0 0 1 5.3 -4.49 c .4 -.12 .81 .57 1.24 1.28 .4 .68 .82 1.37 1.24 1.37 .45 0 .9 -.68 1.33 -1.35 .45 -.7 .89 -1.38 1.32 -1.25 a 8.61 8.61 0 0 1 5 4.17 c 3.73 -2.94 5.1 -7.74 5.1 -10.7 0 -2.34 -1.57 -1.6 -4.09 -.36 l -.14 .07 c -2.31 1.15 -5.39 2.67 -8.77 2.67 s -6.45 -1.52 -8.77 -2.67 c -2.6 -1.29 -4.23 -2.1 -4.23 .29 0 3.05 1.46 8.06 5.47 10.97 Z};\fill[hfFF9D0B] svg {M 70.71 37 a 3.25 3.25 0 1 0 0 -6.5 3.25 3.25 0 0 0 0 6.5 Z M 24.21 37 a 3.25 3.25 0 1 0 0 -6.5 3.25 3.25 0 0 0 0 6.5 Z M 17.52 48 c -1.62 0 -3.06 .66 -4.07 1.87 a 5.97 5.97 0 0 0 -1.33 3.76 7.1 7.1 0 0 0 -1.94 -.3 c -1.55 0 -2.95 .59 -3.94 1.66 a 5.8 5.8 0 0 0 -.8 7 5.3 5.3 0 0 0 -1.79 2.82 c -.24 .9 -.48 2.8 .8 4.74 a 5.22 5.22 0 0 0 -.37 5.02 c 1.02 2.32 3.57 4.14 8.52 6.1 3.07 1.22 5.89 2 5.91 2.01 a 44.33 44.33 0 0 0 10.93 1.6 c 5.86 0 10.05 -1.8 12.46 -5.34 3.88 -5.69 3.33 -10.9 -1.7 -15.92 -2.77 -2.78 -4.62 -6.87 -5 -7.77 -.78 -2.66 -2.84 -5.62 -6.25 -5.62 a 5.7 5.7 0 0 0 -4.6 2.46 c -1 -1.26 -1.98 -2.25 -2.86 -2.82 A 7.4 7.4 0 0 0 17.52 48 Z m 0 4 c .51 0 1.14 .22 1.82 .65 2.14 1.36 6.25 8.43 7.76 11.18 .5 .92 1.37 1.31 2.14 1.31 1.55 0 2.75 -1.53 .15 -3.48 -3.92 -2.93 -2.55 -7.72 -.68 -8.01 .08 -.02 .17 -.02 .24 -.02 1.7 0 2.45 2.93 2.45 2.93 s 2.2 5.52 5.98 9.3 c 3.77 3.77 3.97 6.8 1.22 10.83 -1.88 2.75 -5.47 3.58 -9.16 3.58 -3.81 0 -7.73 -.9 -9.92 -1.46 -.11 -.03 -13.45 -3.8 -11.76 -7 .28 -.54 .75 -.76 1.34 -.76 2.38 0 6.7 3.54 8.57 3.54 .41 0 .7 -.17 .83 -.6 .79 -2.85 -12.06 -4.05 -10.98 -8.17 .2 -.73 .71 -1.02 1.44 -1.02 3.14 0 10.2 5.53 11.68 5.53 .11 0 .2 -.03 .24 -.1 .74 -1.2 .33 -2.04 -4.9 -5.2 -5.21 -3.16 -8.88 -5.06 -6.8 -7.33 .24 -.26 .58 -.38 1 -.38 3.17 0 10.66 6.82 10.66 6.82 s 2.02 2.1 3.25 2.1 c .28 0 .52 -.1 .68 -.38 .86 -1.46 -8.06 -8.22 -8.56 -11.01 -.34 -1.9 .24 -2.85 1.31 -2.85 Z};\fill[hfFFD21E] svg {M 38.6 76.69 c 2.75 -4.04 2.55 -7.07 -1.22 -10.84 -3.78 -3.77 -5.98 -9.3 -5.98 -9.3 s -.82 -3.2 -2.69 -2.9 c -1.87 .3 -3.24 5.08 .68 8.01 3.91 2.93 -.78 4.92 -2.29 2.17 -1.5 -2.75 -5.62 -9.82 -7.76 -11.18 -2.13 -1.35 -3.63 -.6 -3.13 2.2 .5 2.79 9.43 9.55 8.56 11 -.87 1.47 -3.93 -1.71 -3.93 -1.71 s -9.57 -8.71 -11.66 -6.44 c -2.08 2.27 1.59 4.17 6.8 7.33 5.23 3.16 5.64 4 4.9 5.2 -.75 1.2 -12.28 -8.53 -13.36 -4.4 -1.08 4.11 11.77 5.3 10.98 8.15 -.8 2.85 -9.06 -5.38 -10.74 -2.18 -1.7 3.21 11.65 6.98 11.76 7.01 4.3 1.12 15.25 3.49 19.08 -2.12 Z};\fill[hfFF9D0B] svg {M 77.4 48 c 1.62 0 3.07 .66 4.07 1.87 a 5.97 5.97 0 0 1 1.33 3.76 7.1 7.1 0 0 1 1.95 -.3 c 1.55 0 2.95 .59 3.94 1.66 a 5.8 5.8 0 0 1 .8 7 5.3 5.3 0 0 1 1.78 2.82 c .24 .9 .48 2.8 -.8 4.74 a 5.22 5.22 0 0 1 .37 5.02 c -1.02 2.32 -3.57 4.14 -8.51 6.1 -3.08 1.22 -5.9 2 -5.92 2.01 a 44.33 44.33 0 0 1 -10.93 1.6 c -5.86 0 -10.05 -1.8 -12.46 -5.34 -3.88 -5.69 -3.33 -10.9 1.7 -15.92 2.78 -2.78 4.63 -6.87 5.01 -7.77 .78 -2.66 2.83 -5.62 6.24 -5.62 a 5.7 5.7 0 0 1 4.6 2.46 c 1 -1.26 1.98 -2.25 2.87 -2.82 A 7.4 7.4 0 0 1 77.4 48 Z m 0 4 c -.51 0 -1.13 .22 -1.82 .65 -2.13 1.36 -6.25 8.43 -7.76 11.18 a 2.43 2.43 0 0 1 -2.14 1.31 c -1.54 0 -2.75 -1.53 -.14 -3.48 3.91 -2.93 2.54 -7.72 .67 -8.01 a 1.54 1.54 0 0 0 -.24 -.02 c -1.7 0 -2.45 2.93 -2.45 2.93 s -2.2 5.52 -5.97 9.3 c -3.78 3.77 -3.98 6.8 -1.22 10.83 1.87 2.75 5.47 3.58 9.15 3.58 3.82 0 7.73 -.9 9.93 -1.46 .1 -.03 13.45 -3.8 11.76 -7 -.29 -.54 -.75 -.76 -1.34 -.76 -2.38 0 -6.71 3.54 -8.57 3.54 -.42 0 -.71 -.17 -.83 -.6 -.8 -2.85 12.05 -4.05 10.97 -8.17 -.19 -.73 -.7 -1.02 -1.44 -1.02 -3.14 0 -10.2 5.53 -11.68 5.53 -.1 0 -.19 -.03 -.23 -.1 -.74 -1.2 -.34 -2.04 4.88 -5.2 5.23 -3.16 8.9 -5.06 6.8 -7.33 -.23 -.26 -.57 -.38 -.98 -.38 -3.18 0 -10.67 6.82 -10.67 6.82 s -2.02 2.1 -3.24 2.1 a .74 .74 0 0 1 -.68 -.38 c -.87 -1.46 8.05 -8.22 8.55 -11.01 .34 -1.9 -.24 -2.85 -1.31 -2.85 Z};\fill[hfFFD21E] svg {M 56.33 76.69 c -2.75 -4.04 -2.56 -7.07 1.22 -10.84 3.77 -3.77 5.97 -9.3 5.97 -9.3 s .82 -3.2 2.7 -2.9 c 1.86 .3 3.23 5.08 -.68 8.01 -3.92 2.93 .78 4.92 2.28 2.17 1.51 -2.75 5.63 -9.82 7.76 -11.18 2.13 -1.35 3.64 -.6 3.13 2.2 -.5 2.79 -9.42 9.55 -8.55 11 .86 1.47 3.92 -1.71 3.92 -1.71 s 9.58 -8.71 11.66 -6.44 c 2.08 2.27 -1.58 4.17 -6.8 7.33 -5.23 3.16 -5.63 4 -4.9 5.2 .75 1.2 12.28 -8.53 13.36 -4.4 1.08 4.11 -11.76 5.3 -10.97 8.15 .8 2.85 9.05 -5.38 10.74 -2.18 1.69 3.21 -11.65 6.98 -11.76 7.01 -4.31 1.12 -15.26 3.49 -19.08 -2.12 Z};\end{scope}}}}
\newcommand{\termlogo}[3]{{\tikz[baseline={([yshift=-0.5ex]current bounding box.center)}]{%
  \fill[#2,rounded corners=0.2*#1] (0,0) rectangle (1.2*#1,#1);
  \draw[#3,line width=0.11*#1,line cap=round,line join=round] (0.27*#1,0.7*#1) -- (0.5*#1,0.5*#1) -- (0.27*#1,0.3*#1);
  \draw[#3,line width=0.11*#1,line cap=round] (0.6*#1,0.29*#1) -- (0.93*#1,0.29*#1);}}}

\definecolor{ghdark}{HTML}{24292F}
\definecolor{hfpale}{HTML}{FFF8DB}
\definecolor{pyblue}{HTML}{3775A9}
\tcbset{linkbutton/.style={on line,enhanced,arc=9pt,left=8pt,right=9pt,top=4pt,bottom=4pt}}

\newcommand{\eps}{\varepsilon}
\newcommand{\B}{\mathcal{B}}

\newcommand{\method}{Tau-Anchor\xspace}
\newcommand{\clipdet}{CLIP-LP\xspace}
\newcommand{\restrav}{ReStraV\xspace}
\newcommand{\flowts}{Flow-TS\xspace}
\newcommand{\codecmv}{Codec-MV\xspace}

\newcommand{\mvmob}{MV-MobileNet\xspace}
\newcommand{\stream}{STREAM\xspace}
\newcommand{\streammv}{STREAM-MV\xspace}
\newcommand{\gvb}{GenVidBench\xspace}
\newcommand{\aigvd}{AIGVDBench\xspace}
\newcommand{\genvideo}{GenVideo\xspace}
\newcommand{\genbuster}{GenBuster-Bench\xspace}
\newcommand{\vif}{ViF-Bench\xspace}
\newcommand{\dclip}{$\Delta$-CLIP\xspace}

\makeatletter
\newenvironment{llmusage}{\KIL@back{LLM usage statement}}{\par\addvspace{4pt}}
\makeatother

\title{CivicBench: Robust Detection of AI-Generated Political Video}
\RunningTitle{CivicBench: Robust detection of AI-generated political video}

\Author[orcid=0009-0001-0798-1361]{Mert Onur Cakiroglu}{iub}
\Author[orcid=0000-0001-7774-4604]{Elham Buxton}{uis}
\Author[orcid=0000-0001-9091-0240]{Mehmet Dalkilic}{iub}
\Author[corresponding=hkurban@hbku.edu.qa, orcid=0000-0003-3142-2866]{Hasan Kurban}{hbku}

\CodeURL{https://github.com/KurbanIntelligenceLab/CivicBench}

\begin{document}
\maketitle
\vspace{6pt}{\centering
\href{\GHURL}{\tcbox[linkbutton,colback=ghdark,colframe=ghdark,boxrule=0pt,drop fuzzy shadow=black!35]{%
  \ghlogo{11pt}{white}\hspace{0.4em}{\kilsemi\small\color{white}GitHub}}}\hspace{0.8em}%
\href{\HFURL}{\tcbox[linkbutton,colback=hfpale,colframe=hfFFD21E,boxrule=0.8pt,drop fuzzy shadow=black!25]{%
  \hflogo{12.5pt}\hspace{0.4em}{\kilsemi\small\color{hf3A3B45}Hugging Face}}}\hspace{0.8em}%
\href{\PYURL}{\tcbox[linkbutton,colback=pyblue,colframe=pyblue,boxrule=0pt,drop fuzzy shadow=black!30]{%
  \termlogo{10pt}{white}{pyblue}\hspace{0.4em}{\footnotesize\ttfamily\color{white}pip install civicbench}}}\par}\vspace{-6pt}

\begin{abstract}
AI-generated video of politicians can pass for real footage, and once such fakes are common, real footage can be dismissed as fake. Platforms address both problems with detectors, but whoever uploads a clip controls its pixels. Existing datasets of political video are small, show real politicians, and test no attacks. We introduce CivicBench, with 10,800 generated clips of invented politicians from eight open generators and 2,632 public-domain or openly licensed real clips. Each clip can be attacked, to hide a fake or to frame real footage. Published detectors trained on a general benchmark reach at most .781 AUC on CivicBench. Adversarial changes of 1 gray level per pixel hide nearly every generated clip from the detectors not built to resist attacks. An attack's reach is how far it can move a clip's score. A clip keeps its label when its margin from the threshold exceeds this reach, and we find that the reach grows almost linearly with the budget. On this basis, we propose \method, which fine-tunes an encoder pretrained on attacked images. It attacks real and generated clips during training and adds a margin term at its deployed threshold. It reaches .941 AUC on CivicBench. It also keeps .668 of generated clips flagged at 2 gray levels, against at most .289 for published detectors. When real clips are framed at the same 2 gray levels, \method's false alarms rise only from 8\% to 14\%, whereas four published detectors flag nearly every framed real clip at just 1 gray level. External political footage and two general benchmarks show the same pattern under attack. Code and data are available at \url{https://github.com/KurbanIntelligenceLab/CivicBench}.
\end{abstract}

\printkeywords
\begin{figure}[t]
\centering
\includegraphics[width=\linewidth]{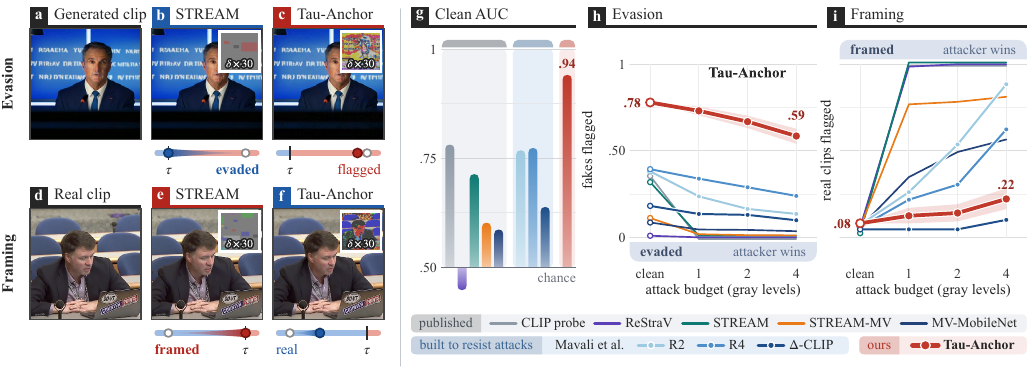}
\caption{\textbf{Detectors of political video, clean and under attack.} (a--c) Evasion of a generated CivicBench clip of an invented senator. The Square attack, which needs only scores, evades \stream at 1 gray level, one step on the 0 to 255 scale of a pixel (b). \method keeps the clip flagged under projected gradient descent (PGD) at 2 gray levels (c). (d--f) Framing of a real State Department clip. \stream flags it at 1 gray level (e), and \method does not at 2 (f). Insets show the perturbation, magnified 30 times. Gauges show the clean score (open), the attacked score (filled), and the threshold $\tau$. (g) Clean area under the ROC curve (AUC) on CivicBench, with the detectors in the order of the legend and a lid in the color of each group. (h) Share of generated clips that stay flagged under evasion and (i) share of real clips flagged under framing, without an attack (clean) and at 1, 2, and 4 gray levels. The band is \method's 95\% CI.}
\label{fig:teaser}
\end{figure}

\section{Introduction}
\label{sec:intro}
Generated videos of politicians now appear around votes and wars. In 2022, a fake Zelenskyy told Ukrainian soldiers to lay down their arms~\cite{npr2022zelensky}, and fake videos have since announced that Moldova's president had resigned and that a candidate in Ireland had withdrawn~\cite{balkaninsight2023sandu,rte2025connolly}. Such videos do harm in two ways. First, a fake can pass as a real statement. Second, once fakes are common, real footage can be dismissed as fake, an effect known as the liar's dividend~\cite{chesney2019deep}. In 2025, the US president called a video from the White House AI-generated after his staff had said it was real~\cite{ap2025whitehouse}. The EU now requires generated videos to be marked so that they can be detected~\cite{euaiact2024}. Some platforms label such clips, though not consistently~\cite{rte2025connolly}.

These efforts highlight the need to distinguish fake videos from real footage. Yet even automated detectors are vulnerable to manipulation. Whoever uploads a clip can make small changes to its pixels to steer a detector's decision. An attacker can alter a fake video so that the detector classifies it as real. We call this evasion. Conversely, an attacker can alter real footage so that the detector flags it as fake. We call this framing (\cref{fig:teaser}).

Existing data cannot measure this risk for political video. A benchmark built from real political deepfakes holds 146 deepfake videos~\cite{lin2025fit}. Deepfakes collected from the web~\cite{chandra2025deepfakeeval} and recent benchmarks of generated video~\cite{shykula2026df26,liang2026crisis} include politics only as a small part. Because these datasets show real, identifiable politicians, their release is limited. To our knowledge, none of these datasets or benchmarks are designed to test adaptive attacks that hide a fake video or frame a real clip. To fill this gap, we build CivicBench (\cref{sec:civic}). It includes 10,800 generated clips of invented politicians, parties, and countries in nine kinds of civic scenes, generated by eight open generators from the same 1,350 prompts. We use open generators deliberately because they carry no mark that a detector could rely on and are therefore more likely to be used by attackers. CivicBench also includes 2,632 real clips of public-domain and openly licensed footage of briefings, speeches, legislatures, and protests. CivicBench supports both attack directions (evasion and framing).

We test published detectors that were trained on a general video benchmark, without retraining them on political video. Even without an attack, they perform poorly on CivicBench, with an area under the ROC curve (AUC) of only .448 to .781 (\cref{fig:teaser}g). They also flag two to three times as much real political footage as intended: 10.6\% to 15.1\%, against a target of 5\% set during training (\cref{sec:main}). Under attack, they perform even worse. Adversarial changes of 1 gray level, that is, at most 1/255 per pixel, hide nearly every generated clip from the detectors not built to resist attacks. They also make four of these detectors flag nearly every attacked real clip (\cref{tab:main}). Detectors built to resist attacks~\cite{mavali2026adversarial,shahjahan2026improving} do better, but still keep at most .289 of generated clips flagged at 2 gray levels (\cref{fig:teaser}h). They were designed for images, not video, and like adversarial training in general~\cite{madry2018towards}, they are rarely tested against framing. Detectors based on codec motion vectors can appear robust because the codec is not differentiable, but we show that small perturbations change many motion vectors, which makes these detectors vulnerable as well (\cref{sec:mv}).

These failures share an underlying condition. A detector flags a clip when its score reaches a threshold $\tau$, set for a fixed false-alarm rate. An attack's reach is how far it can move a clip's score. A clip keeps its label under attack when its margin from $\tau$ exceeds this reach (\cref{sec:margin}). A detector can therefore become more robust in two ways: (1) by reducing the reach, which depends mainly on the encoder, and (2) by increasing the margin of the fake videos nearest to $\tau$, which are the first to be misclassified under attack. We propose \method, a detector that combines both approaches at the threshold at which it is deployed. To reduce the reach, \method builds on an image encoder that was pretrained to resist adversarial perturbations. It then fine-tunes this encoder on attacked real and generated clips from \aigvd~\cite{ma2026your}. To increase the margin, \method adds a loss term that pushes the fake videos nearest to $\tau$ further above it.

Our contributions are:
\begin{itemize}[leftmargin=*,itemsep=1pt,topsep=2pt]
\item \textbf{CivicBench}, to our knowledge the first benchmark that tests detectors of political video against adaptive attacks in both directions. Its politicians are invented, so its generated clips can be shared. We release it on Hugging Face, with a pip package that runs its attacks on any detector.
\item \textbf{\method}, a video detector built on an adversarially pretrained encoder and trained at its deployed threshold. It reaches an AUC of .941 on CivicBench and keeps .668 of political fakes flagged at 2 gray levels, against at most .289 for published detectors.
\item \textbf{Margin against reach}, a condition that determines whether a clip keeps its label under attack. Since the reach grows almost linearly in the budget, an attack of 1 gray level predicts the share of clips that keep their label at 4 gray levels to within 11 percentage points. A bound on codec motion vectors explains why detectors that read them fail.
\item \textbf{A two-sided evaluation} on CivicBench and external political footage, with evasion also on two general benchmarks, showing that tests of evasion alone miss attacks that turn a detector against real footage.
\end{itemize}

\section{Related Work}
\label{sec:related}

\begin{figure}[t]
\centering
\includegraphics[width=\linewidth]{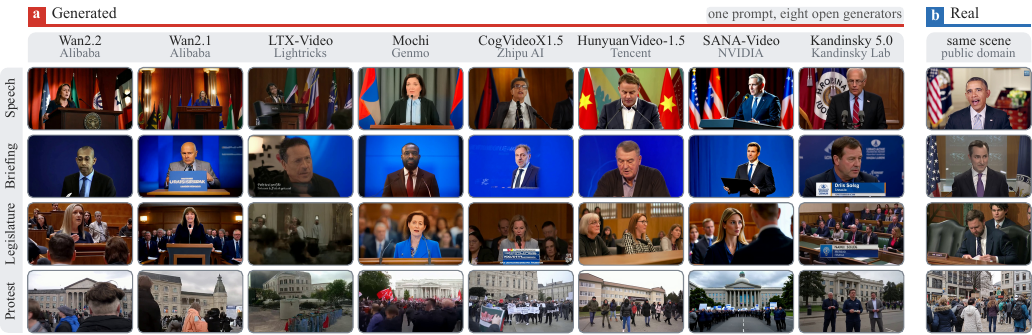}
\caption{\textbf{CivicBench.} (a) The middle frame of one prompt per kind of scene, rendered by each of the eight generators. In the legislature prompt, Naomi Solberg addresses parliament in Norvenia. She is an invented opposition leader of the Meridian Front. (b) A real clip of the same kind of scene, from public-domain footage.}
\label{fig:gallery}
\end{figure}

\paragraph{Political deepfakes} Generated video has targeted elections, from fakes of film stars attacking India's prime minister~\cite{reuters2024india} to a revived Suharto before Indonesia's vote~\cite{cnn2024suharto}. Datasets of real incidents~\cite{lin2025fit,chandra2025deepfakeeval} and benchmarks with news and crisis scenes~\cite{shykula2026df26,liang2026crisis} show real people and measure accuracy without an attacker. CivicBench uses invented people, pairs every prompt with eight generators, and attacks both generated and real clips.

\paragraph{Detecting AI-generated video} Frame-level detectors read features of pretrained encoders such as CLIP~\cite{radford2021learning,ojha2023towards} or DINOv2~\cite{oquab2024dinov2}, sometimes with temporal statistics of their trajectories~\cite{interno2025ai}. Universal detectors pool frame and background cues~\cite{kundu2025towards}. Compressed-domain detectors read codec motion vectors and residuals~\cite{zhu2026beyond,cakiroglu2026detect}. Benchmarks now cover dozens of generators~\cite{ma2026your,ni2026genvidbench,chen2024demamba,wen2025busterx}. These works measure accuracy on unaltered, compressed, or resized clips, not under adaptive attacks.

\paragraph{Robust detection and adversarial training} Small perturbations evade image forensic detectors~\cite{carlini2020evading,hussain2022exposing,saberi2024robustness}. Two recent detectors with public code are built to resist attacks. Mavali et al.~\cite{mavali2026adversarial} train a linear probe on an adversarially robust CLIP encoder, and Shahjahan et al.~\cite{shahjahan2026improving} combine CLIP features with a denoising residual. Neither addresses video, framing, or the deployed threshold. Adversarial training~\cite{madry2018towards}, TRADES~\cite{zhang2019theoretically}, and MART~\cite{wang2020improving} remain the standard defenses. Max-margin training enlarges each input's distance to the decision boundary~\cite{ding2020mma}, whereas \method's margin acts at a fixed threshold. Averaging weights reduces the decline of robustness late in training~\cite{rice2020overfitting,rebuffi2021data}. Robust CLIP encoders~\cite{mao2023understanding,schlarmann2024robust,wang2025double} transfer robustness to linear heads~\cite{nern2023transfer}. \method brings these parts to video. What it adds is a margin at the deployed threshold, and we test it against attacks in both directions.

\section{CivicBench}
\label{sec:civic}

CivicBench pairs generated clips of invented politicians with real political footage (\cref{fig:gallery}). We use it exclusively for evaluation and do not train any detectors on its clips.

\paragraph{Prompts} Templates describe nine kinds of civic scenes: speeches, press briefings, legislatures, rallies, protests, newsrooms, interviews, debates, and crisis reports. They are the scenes in which documented political deepfakes pose as footage. In the Political Deepfakes Incidents Database (PDID)~\cite{walker2024merging}, about 39\% of the 1,059 political deepfake videos show one of them, most often a speech (\cref{supp:civic}). Each prompt fills its scene with invented names, parties, countries, and news channels. Each prompt has one of three levels of political content. Level 0 shows the scene with no political words, level 1 names an invented official and role, and level 2 adds an election, such as a victory speech. No template describes voter suppression, ballot tampering, or violence.

\paragraph{Generators} Eight open text-to-video models from seven groups render every prompt (\cref{tab:civic_gen}). Shared prompts make the generators comparable clip by clip. Commercial services mark their video~\cite{google2024synthidveo,openai2025sora2card} and block many requests for real politicians~\cite{openai2025sora2card,chow2025veo3time}. Open models, in contrast, run on one consumer GPU~\cite{wan2025github}. They are also easy to misuse. The watermark of an open image generator can be removed by editing one line of code~\cite{fernandez2023stablesignature}. Within about a day of its release, one open video model was adopted by a community that makes intimate videos of real people without consent~\cite{maiberg2025wan}. CivicBench therefore uses open generators. Closed generators such as Sora~2 and Veo~3.1 appear in the Wild split of \genbuster~\cite{wen2025busterx}, which we test in \cref{supp:genvideo}.

\begin{table}[t]
\centering\footnotesize
\caption{\textbf{Generators of CivicBench.} Each renders the same 1,350 prompts (settings in \cref{supp:civic}).}
\label{tab:civic_gen}
\setlength{\tabcolsep}{2.4pt}
\begin{tabular}{@{}llrrr@{}}
\toprule
Generator & Group & Size & Frames@fps & Pixels\\
\midrule
Wan2.2-TI2V~\cite{wan2025wan} & Alibaba & 5B & 49 @ 24 & 1280$\times$704\\
Wan2.1-T2V~\cite{wan2025wan} & Alibaba & 1.3B & 33 @ 16 & 832$\times$480\\
LTX-Video~\cite{hacohen2024ltxvideo} & Lightricks & 2B & 49 @ 24 & 768$\times$512\\
Mochi 1~\cite{genmo2024mochi} & Genmo & 10B & 43 @ 30 & 848$\times$480\\
CogVideoX1.5~\cite{yang2025cogvideox} & Zhipu AI & 5B & 33 @ 16 & 1360$\times$768\\
HunyuanVideo-1.5~\cite{wu2025hunyuanvideo15} & Tencent & 8.3B & 61 @ 24 & 848$\times$480\\
SANA-Video~\cite{chen2025sanavideo} & NVIDIA & 2B & 81 @ 16 & 832$\times$480\\
Kandinsky 5.0~\cite{arkhipkin2026kandinsky5} & Kandinsky Lab & 2B & 61 @ 24 & 768$\times$512\\
\bottomrule
\end{tabular}
\end{table}

\paragraph{Real footage} The real clips come from the White House, the State Department, the US Senate, the Kremlin, and recordings of demonstrations. All are public domain or openly licensed (\cref{tab:civic_sources}). The 2,632 clips show speeches (891), protests (806), briefings (599), and legislatures (336). We use these four kinds for two reasons. First, they make up 74\% of the PDID videos that show one of the nine kinds, and speeches alone make up 65\%. Second, official channels publish them under open licenses, whereas broadcast news, interviews, and debates are mostly copyrighted. Because the generated clips include the same four kinds, the two classes can also be compared on matched scenes. Each kind comes mostly from one source, so a detector could learn the source instead of the class. Leaving any one source out keeps \method's AUC between .933 and .949 (\cref{supp:civic}).

\paragraph{Shortcuts} Real and generated clips come from different sources, so a detector could separate them by how a clip was stored rather than by what it shows. Our shared processing (\cref{supp:setup}) removes clip length and frame size, but not the sampling rate. Generated clips are sampled at 8 frames per second (Mochi's at 7.5), and real ones at 6.0 to 10.0. On real clips sampled at the same rate as the generated ones, \method still reaches an AUC of .957, against .611 to .812 for the other detectors. Real and generated clips also differ before processing, in resolution and in how often they were compressed. We downscaled 924 real clips to 480 pixels and encoded them once with H.264, as a generated clip would be. This changes no detector's AUC by more than .05 (\cref{supp:civic}).

\paragraph{Release} Real and generated clips are hosted openly on Hugging Face. Real clips keep their own licenses, with credits and source links, and generated clips are released for research on detection (\cref{supp:civic}).

\section{Why Detectors Fail under Attack}
\label{sec:theory}

This section shows when a clip keeps its label under attack and why detectors that read codec motion vectors lose it. Both results set the requirements for \method.

\subsection{Detection at an operating point}
\label{sec:setup0}

A detector gives a clip $x$ of $T$ frames with 8-bit pixels a score $s(x)$ and flags it as generated when $s(x)\geq\tau$. A fact-checker can follow up only a few false alarms, so we set $\tau$ for a false-alarm rate of 5\%, the share of real clips that are flagged. The true-positive rate (TPR) is the share of generated clips that are flagged. An attacker changes a clip by a perturbation $\delta$ small enough to be hard to see. The budget $\eps$ bounds the change of each pixel in gray levels, $\norm{\delta}_\infty\leq\eps$. An attacker can turn $x$ into any 8-bit clip that differs from it by at most $\eps$ gray levels in every pixel. We write $\B_\eps(x)$ for the set of all such clips, the clips reachable from $x$. The attacker has two goals. (i)~To hide a fake, evasion succeeds on a flagged generated clip $x$ when some clip in $\B_\eps(x)$ is not flagged. The TPR under evasion is the share of generated clips that stay flagged on all of $\B_\eps(x)$. (ii)~To discredit real footage, framing succeeds on an unflagged real clip $x$ when some clip in $\B_\eps(x)$ is flagged. The attacked false-alarm rate is the share of real clips flagged after the attack.

\subsection{Margin against reach}
\label{sec:margin}

Many detectors embed a clip as a vector $e(x)$ and score it with a linear head, $s(x)=w^\top e(x)+b$. The margin of a clip is its signed distance from the threshold along $w$, positive on the side of its label: $\rho(x)=(s(x)-\tau)/\norm{w}$ for a generated clip and $\rho(x)=(\tau-s(x))/\norm{w}$ for a real one. The reach of an attack is the farthest it can move the embedding toward the threshold along $w$,
\begin{equation}
  r(x,\eps)=\max_{x'\in\B_\eps(x)} \frac{\pm\,w^\top\big(e(x)-e(x')\big)}{\norm{w}},
  \label{eq:disp}
\end{equation}
with $+$ for a generated clip and $-$ for a real one, measured in embedding units.

\begin{figure}[t]
\centering
\includegraphics[width=0.55\linewidth]{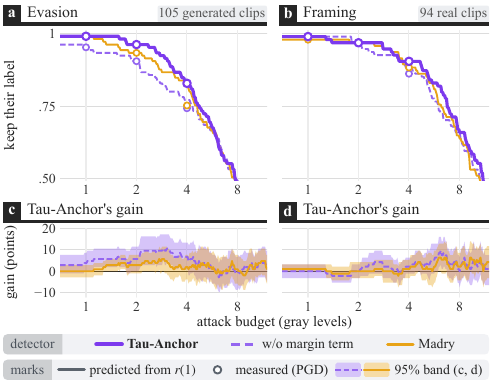}
\caption{\textbf{Margin against reach on CivicBench} (\cref{prop:survival}). (a, b)~Share of clips that keep their label, among the 105 generated and 94 real clips attacked against \method, \method without its margin term, and Madry's objective. Lines are predicted from each clip's margin and reach at 1 gray level, circles measured with PGD. Every \stream clip breaks below 0.8 gray levels, less than the smallest change of a pixel (not shown). (c, d)~\method's gain over each other detector, in percentage points, with 95\% bootstrap bands.}
\label{fig:reach}
\end{figure}

\begin{proposition}[Keeping the label]
\label{prop:survival}
A flagged generated clip $x$ stays flagged for every clip in $\B_\eps(x)$ if and only if $\rho(x)\geq r(x,\eps)$, and an unflagged real clip stays unflagged if and only if $\rho(x)>r(x,\eps)$. The TPR under evasion at budget $\eps$ is therefore the probability, over generated clips, that $\rho(x)\geq r(x,\eps)$, and the attacked false-alarm rate is the probability, over real clips, that $\rho(x)\leq r(x,\eps)$.
\end{proposition}

The condition is the threshold test applied to the worst reachable clip, as in margin certificates~\cite{hein2017formal,tsuzuku2018lipschitz} (proof in \cref{supp:proofs}). Stated in score units, the condition holds for any detector, with or without a linear head. We measure the reach with projected gradient descent (PGD)~\cite{madry2018towards}, which finds one reachable clip, so the measured reach is a lower bound. Robustness thus depends on two quantities, the margin of the clips near the threshold and the reach of the attack. On CivicBench the reach grows almost exactly linearly in the budget. From 1 to 4 gray levels, it grows by a factor of 3.8 to 4.4 for the detectors on the robust encoder \dclip~\cite{wang2025double}. A clip therefore keeps its label up to about $\rho(x)/r(x,1)$ gray levels, which we call its break budget. As a result, one attack of 1 gray level predicts the share of clips that break at 4 to within 11 percentage points, for evasion and framing alike (\cref{fig:reach}a, b).

Of the two, the reach matters more. It depends mainly on the encoder (\cref{supp:theory}). On CivicBench, the median break budget under evasion is 0.08 gray levels for \stream, below the smallest change of a pixel. It is 5.3 to 8.4 for the detectors on the robust encoder \dclip.

\subsection{Motion vectors flip under the smallest attack}
\label{sec:mv}

A compressed video stores most frames as changes to the previous one. For each block of pixels, the codec stores a motion vector, the integer-pixel offset to the best-matching block of the previous frame, its reference. Some detectors read these vectors as a summary of motion, and they can look robust because gradients do not pass through the codec. The best match has the lowest cost, the mean absolute difference between the two blocks. Let $g$ be the gap between the lowest and second-lowest cost of a block.

\begin{proposition}[Near ties]
\label{prop:ties}
Perturbing both frames by at most $\eps$ per pixel changes every cost by at most $2\eps$, so a motion vector stays unchanged whenever $g>4\eps$. When only the current frame may change, a block with $g<2\eps q$ can be flipped, where $q$ is the share of its pixels whose value lies between those of its two best references, at least $\eps$ from the second.
\end{proposition}

In practice the gaps are small, with a median of 0.15 to 0.34 gray levels. Only 5\% to 11\% of blocks are guaranteed stable at 1 gray level, in real and generated clips alike (\cref{supp:mvties}). The smallest attack can thus rewrite many motion vectors, so detectors that read them have a large reach in the sense of \cref{prop:survival}. The motion field is also piecewise constant, so its gradient is zero almost everywhere, and we attack these detectors with searches through the actual codec, not only with gradients through smooth stand-ins (\cref{supp:mvties}).

\subsection{Requirements}
\label{sec:requirements}

\Cref{prop:survival,prop:ties} set two requirements for a detector that keeps its labels under attack. \textbf{(I)} The reach must stay below the margin of the clips near the threshold, which calls for an encoder made robust by pretraining, not for motion vectors. \textbf{(II)} The fakes nearest the deployed threshold break first, so training should enlarge their margin at that threshold.

\section{\method}
\label{sec:method}

\paragraph{Encoder (I)} \method embeds each of the $T=11$ frames of a clip with \dclip~\cite{wang2025double}, a ViT-H/14 CLIP encoder pretrained at scale on attacked images. It normalizes the frame embeddings, averages them, and normalizes the mean to obtain $e(x)$, which a linear head scores as $s(x)=w^\top e(x)+b$ (\cref{fig:arch}). We fine-tune the last four transformer blocks and the head, starting the head from a linear probe.

\paragraph{Attacks on both classes} Each training step attacks the batch as fast adversarial training does~\cite{wong2020fast}. The attack starts from a random perturbation within 4 gray levels and takes one step of 2 gray levels along the sign of the loss gradient. Generated clips are pushed toward real, and real clips toward generated. These are the two directions of \cref{prop:survival}.

\paragraph{Margin at the deployed threshold (II)} By \cref{prop:survival}, the fakes just above the deployed threshold break first. Standard adversarial objectives move the attacked score $z'=s(x+\delta)$ without reference to the threshold. We add a margin term with weight $\gamma$ and target margin $\kappa$ on the attacked fakes,
\begin{equation}
  \mathcal{L}_\rho=\gamma\,\mathrm{softplus}\!\left(\frac{\kappa-\rho(x+\delta)}{\kappa}\right),
  \label{eq:margin}
\end{equation}
where $\mathrm{softplus}(u)=\log(1+e^u)$ and $\rho$ is measured at a running estimate $\hat\tau$ of the deployed threshold, the 95th percentile of the scores of real training clips.

\begin{proposition}[Push at the deployed threshold]
\label{prop:align}
Let $z=s(x)$ and $z'=s(x+\delta)$ be the logits of a generated clip and of its attacked version, and call the size of a loss's gradient on a score its push. (i)~Cross-entropy on the attacked clip pushes $z'$ up by $\sigma(-z')$, where $\sigma$ is the logistic function. (ii)~The TRADES term $\eta\,\mathrm{KL}\big(\sigma(z)\,\|\,\sigma(z')\big)$ pushes $z'$ toward $z$ by $\eta|\sigma(z)-\sigma(z')|$, which is zero whenever $z'=z$, for any $z$. (iii)~$\mathcal{L}_\rho$ pushes $\rho(x+\delta)$ up by at least $\gamma/(2\kappa)$ for every attacked fake with $\rho(x+\delta)\leq\kappa$, for any $\hat\tau$.
\end{proposition}

Cross-entropy thus mainly pushes fakes whose logit is below 0. The TRADES term only resists changes of the logit. Only the margin term targets a margin above the threshold.

\paragraph{Full objective} The loss is the sum of cross-entropy on clean and attacked clips, the margin term of \cref{eq:margin}, and a preservation term $\lambda\norm{e(x)-e_0(x)}^2$ that keeps the embedding near that of the original encoder $e_0$. Cross-entropy on fakes is weighted per generator, more for generators whose clips sit near the threshold (\cref{supp:setup}). We deploy an exponential moving average of \method's weights with decay 0.999. On \aigvd, this average keeps more attacked fakes flagged at 1 and 2 gray levels (\cref{supp:ci}).

\section{Experiments}
\label{sec:experiments}

\subsection{Setup}
\label{sec:setup}

\paragraph{Training and test data} Detectors are trained on subsets of two general benchmarks. We use 14,980 clips of \aigvd~\cite{ma2026your} from six of its generators and 15,000 clips of \gvb~\cite{ni2026genvidbench} from four, each with real videos. We split each into five folds grouped by prompt, and each fold is tested by a model trained on the other four. On CivicBench, we test the models trained on fold 0 of \aigvd, because the generators of \aigvd are closer to those of CivicBench than the generators of \gvb. Models trained on \gvb appear in \cref{supp:civic}. On every test set, the threshold is set on its own real clips at 5\% false alarms~\cite{carlini2022membership,an2024waves}.

\paragraph{Attacks} The attacker knows the full detector and attacks it end to end. We run PGD (100 steps), Auto-PGD (APGD, 100 steps and 5 restarts)~\cite{croce2020reliable}, and the Square attack (1,000 queries)~\cite{andriushchenko2020square}, which needs only scores. Published detectors and those built to resist attacks face all three. \method, Madry's objective, and the ablations face PGD on all their attacked clips. \method and Madry's objective also face APGD at 4 gray levels on all of them except Mochi's, and Square on 20 clips per direction. Budgets are 1, 2, and 4 gray levels on CivicBench. These changes are small. Even at 4 gray levels, successful attacks have a median peak signal-to-noise ratio (PSNR) of 37.2\,dB. They also match the training budgets of robust CLIP encoders~\cite{schlarmann2024robust}. On the general benchmarks, budgets go up to 8. An attack counts only if the saved 8-bit clip crosses the threshold. Per clip, we keep the attack that succeeds at the lowest budget. On CivicBench, \method and Madry's objective are attacked on the same 225 flagged generated clips and 185 real clips, drawn from all four kinds of real scenes. The ablations and Mavali et al.\ on \dclip are attacked on the first 120 of each. The other detectors are attacked on up to 10 of their own flagged clips per generator and on 69 real clips. On the general benchmarks, we attack 40 flagged generated clips per benchmark and fold. A real clip that a detector flags before the attack counts as framed. Because only flagged fakes are attacked, the TPR under evasion is the clean TPR times the share of attacked fakes that stay flagged.

\paragraph{Methods compared} The six published detectors read different signals. A CLIP probe~\cite{ojha2023towards} on ViT-L/14 and \restrav~\cite{interno2025ai} read frames. D3~\cite{zheng2025d3} reads the change between frame features of a video encoder, without training. \stream reads the encoded stream, and \streammv only its motion vectors~\cite{zhu2026beyond}. \mvmob~\cite{gronquist2024efficient} reads codec motion vectors. Three detectors of a recent audit toolkit~\cite{cakiroglu2026auditing} appear in \cref{supp:attacks}. The detectors built to resist attacks are those of Mavali et al.~\cite{mavali2026adversarial} and Shahjahan et al.~\cite{shahjahan2026improving}. We run the detector of Mavali et al.\ on \dclip and on R2 and R4, CLIP encoders fine-tuned adversarially at 2 and 4 gray levels~\cite{schlarmann2024robust}. Madry's objective, TRADES, and MART replace \method's loss. They keep its encoder, fine-tuning, training attack, and weight averaging. TRADES and MART also drop its generator weights. Every method except D3 is trained on our folds (\cref{supp:setup}).

\paragraph{Statistics} We report 95\% confidence intervals (CIs) from a bootstrap with 2,000 resamples, real and generated clips resampled separately. On CivicBench, clean intervals resample prompts and source videos. For detectors attacked on the same clips, a paired bootstrap resamples those clips. On the general benchmarks, Madry's objective, TRADES, MART, and Mavali et al.\ on R4 and \dclip are attacked on \method's clips over five folds for paired comparisons. The other detectors are attacked on their own flagged clips of fold 0 (\cref{supp:ci}).

\begin{table}[t]
\centering
\caption{\textbf{Detectors under attack.} On CivicBench (left), clean AUC and TPR, then TPR under evasion and attacked false-alarm rate under framing at 1, 2, and 4 gray levels. On the general benchmarks (right), clean AUC and TPR over five folds and TPR under evasion at 2 and 4 gray levels, over five folds for \method and for Mavali et al.\ on R4 and \dclip, and on fold 0 for the others. D3's evasion cells rest on 19 clips. Those of \streammv, \mvmob, and Mavali et al.\ on R2 and R4 are upper bounds (\cref{supp:setup}). $^{\dagger}$Not significantly different from \method (95\% bootstrap). $^\ddagger$Fold 0 only. Its scores change with the GPU model (\cref{supp:setup}).}
\label{tab:main}
\scriptsize
\setlength{\tabcolsep}{2.1pt}
\begin{tabular}{l cc ccc ccc c cccc c cccc}
\toprule
& \multicolumn{8}{c}{CivicBench (political)} && \multicolumn{4}{c}{\aigvd} && \multicolumn{4}{c}{\gvb}\\
\cmidrule{2-9}\cmidrule{11-14}\cmidrule{16-19}
& \multicolumn{2}{c}{clean} & \multicolumn{3}{c}{evasion $\uparrow$} & \multicolumn{3}{c}{framing $\downarrow$} &&
\multicolumn{2}{c}{clean} & \multicolumn{2}{c}{attacked} && \multicolumn{2}{c}{clean} & \multicolumn{2}{c}{attacked}\\
\cmidrule(lr){2-3}\cmidrule(lr){4-6}\cmidrule(lr){7-9}\cmidrule(lr){11-12}\cmidrule(lr){13-14}\cmidrule(lr){16-17}\cmidrule(lr){18-19}
Method & AUC & TPR & 1 & 2 & 4 & 1 & 2 & 4 && AUC & TPR & 2 & 4 && AUC & TPR & 2 & 4\\
\midrule
\multicolumn{19}{l}{\textit{Published detectors}}\\
CLIP probe~\cite{ojha2023towards} & .781 & .364 & .000 & .000 & .000 & 1.000 & 1.000 & 1.000 && .956 & .776 & .000 & .000 && .997 & .989 & .000 & .000\\
\restrav~\cite{interno2025ai} & .448 & .009 & .000 & .000 & .000 & .986 & 1.000 & 1.000 && .646 & .145 & .000 & .000 && .928 & .714 & .000 & .000\\
D3~\cite{zheng2025d3} & .475 & .028 & .000 & .000 & .000 & 1.000 & 1.000 & 1.000 && .619 & .178 & .000 & .000 && .588 & .080 & .000 & .000\\
\stream~\cite{zhu2026beyond} & .714 & .309 & .000 & .000 & .000 & 1.000 & 1.000 & 1.000 && .985 & .930$^{\dagger}$ & .000 & .000 && .995 & .979 & .000 & .000\\
\streammv~\cite{zhu2026beyond} & .602 & .111 & .018 & .013 & .011 & .768 & .783 & .812 && .824 & .358 & .000 & .000 && .950 & .780 & .073 & .023\\
\mvmob~\cite{gronquist2024efficient} & .587 & .086 & .045 & .043 & .035 & .348 & .493 & .565 && .904 & .569 & .057 & .051 && .988 & .956 & .720 & .711\\
\midrule
\multicolumn{19}{l}{\textit{Detectors built to resist attacks}}\\
Mavali et al.~\cite{mavali2026adversarial}, R2 & .768 & .381 & .236 & .164 & .135 & .261 & .536 & .884 && .953 & .777 & .383 & .141 && .992 & .966 & .817 & .312\\
Mavali et al.~\cite{mavali2026adversarial}, R4 & .774 & .394 & .339 & .289 & .239 & .217 & .304 & .623 && .952 & .785 & .584 & .335 && .988 & .949 & .852 & .663\\
Mavali et al.~\cite{mavali2026adversarial}, \dclip & .638 & .182 & .135 & .130 & .099 & \textbf{.046} & \textbf{.046} & \textbf{.101} && .941 & .717 & .628 & .574 && .989 & .943 & .933 & .865\\
Shahjahan et al.~\cite{shahjahan2026improving}$^\ddagger$ & .709 & .233 & .023 & .000 & .000 & .913 & .971 & .986 && \textbf{.997} & \textbf{1.000} & .000 & .000 && .704 & .222 & .000 & .000\\
\midrule
\multicolumn{19}{l}{\textit{Ours}}\\
\method & \textbf{.941} & \textbf{.779} & \textbf{.731} & \textbf{.668} & \textbf{.585} & .124 & .141 & .222 && .988 & .935 & \textbf{.931} & \textbf{.883} && \textbf{.999} & \textbf{.998} & \textbf{.993} & \textbf{.993}\\
\bottomrule
\end{tabular}
\end{table}

\subsection{Main results}
\label{sec:main}

\Cref{tab:main} compares all detectors on CivicBench and on the two general benchmarks. There, too, \method keeps the most attacked fakes flagged (\cref{supp:ci}).

\paragraph{Political video, clean} Without an attack, political video is hard for published detectors, and \method leads them. Applied unchanged, they reach an AUC of .448 to .781 on CivicBench. The CLIP probe transfers best. The detectors that read motion or the change between frames fall furthest. \restrav and D3 even score real clips as more likely generated than the fakes (AUC .448 and .475), partly because their temporal features respond to the frame rate (\cref{supp:civic}). \method detects .779 of the generated clips at 5\% false alarms, against at most .394 for published detectors. Its lead holds on the four kinds of scenes that real and generated clips share (AUC .953, at most .808 for published detectors). It also holds at the three levels of political content and on generators unseen in training (\cref{sec:depth}). Madry's objective on the same encoder is about as accurate (\cref{tab:ablation}). No published detector exceeds an AUC of .805, whichever general benchmark it is trained on (\cref{supp:civic}).

\paragraph{False alarms on real footage} Published detectors flag real political footage even without an attacker. A deployed detector keeps the threshold set in training. At that threshold, the published detectors flag 10.6\% to 15.1\% of the real political clips. Of those from the State Department, they flag 15.9\% to 25.9\%. Mavali et al.\ on \dclip flag 35.1\%, and \method 6.4\%, close to its target of 5\% (\cref{supp:civic}). Framing makes such false alarms deliberate. Below, every threshold is set on CivicBench's own real clips.

\paragraph{Evasion} A budget of 1 gray level hides the generated clips of CivicBench from all six published detectors that we attacked, which keep at most .045 of them flagged. The detectors built to resist attacks keep at most .289 at 2 gray levels. Of the fakes that each of them flags, they keep at most .734, and \method keeps .858. \method keeps .668 and .585 at 2 and 4 gray levels. Madry's objective on the same encoder is on par at 1 and 2 gray levels. On the 179 attacked fakes that both flag, it loses 10.6 percentage points more of them at 4 ($p<.001$). The 95\% CI is [+5.6, +15.6].

\paragraph{Framing} A budget of 1 gray level makes four published detectors flag nearly every attacked real clip, and every other published detector at least 35\% of them. The detectors of Mavali et al.\ are framed less often at 1 gray level but on up to .884 of real clips at 4. On \dclip their detector is framed least, .101 at 4 gray levels against .211 for \method on the same 120 real clips. On their 109 shared clips, the paired difference is $-$11.0 percentage points ($p=.006$). This low rate comes with low sensitivity. CivicBench's 5\% threshold lies far above this detector's deployed threshold, so at the 5\% threshold it detects only .182 of the fakes. At its deployed threshold, it flags 35.1\% of the real clips before any attack. After framing at 2 gray levels, \method flags 14\% of real clips, up from 8\% before the attack. The rate before the attack exceeds 5\% because the attacked real clips overrepresent legislatures, where false alarms are most common (\cref{fig:depth}b). This is on par with Madry's objective on the same clips.

\paragraph{Stronger attacks} Stronger attacks find few evasions that PGD misses. A defense can look robust when the attack is weak or its gradient is masked~\cite{athalye2018obfuscated,carlini2019evaluating,tramer2020adaptive}. On \method's 196 paired attacked fakes, APGD and Square break 1 beyond PGD, at 2 gray levels. They frame no real clip that PGD does not. On fold 0 of the general benchmarks, they evade 4 of 79 attacked clips at a lower budget than PGD, all at 8 gray levels, and PGD with 1,000 steps finds no new evasion at 4 gray levels (\cref{supp:attacks}). An attacker might instead spend a larger budget on a few frames, because \method averages the frame embeddings. On fold 0 of \aigvd, PGD on three frames evades only 8\% of \method's attacked clips, even at 32 gray levels.

\subsection{CivicBench in depth}
\label{sec:depth}

\begin{figure}[t]
\centering
\includegraphics[width=\linewidth]{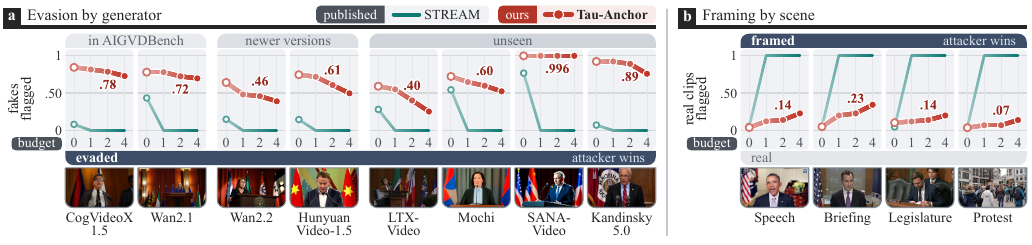}
\caption{\textbf{CivicBench by generator and scene.} (a) TPR under evasion by generator and (b) false-alarm rate under framing by kind of scene, at budgets of 0 (no attack), 1, 2, and 4 gray levels. Numbers give \method at 2 gray levels. The generators are in \aigvd, newer versions of its generators, or unseen. Frames show the same prompt from each generator and a real clip of each scene.}
\label{fig:depth}
\end{figure}

\paragraph{Generators and scenes} \method stays accurate on unseen generators, while attacks break \stream on every generator and kind of scene (\cref{fig:depth}). \method's clean TPR ranges from .590 on LTX-Video to .996 on SANA-Video. Two CivicBench generators, CogVideoX1.5 and Wan2.1, also generate part of \aigvd. Wan2.2 and HunyuanVideo-1.5 are newer versions of its generators. \method's AUC is .946 on the first two and .906 on the newer versions. On the four unseen generators, it is .956. Under evasion at 2 gray levels, its TPR ranges from .401 on LTX-Video to .996 on SANA-Video, while \stream keeps none on any generator (\cref{fig:depth}a).

\paragraph{Uploads} Most attacks survive the H.264 re-encode of an upload at its default quality. This holds for 74\% of the evasions against \stream and 84\% of those that break \method. A platform transcode to 360 pixels undoes more of them. An attack through the transcode frames no real clip against \method beyond those that the transcode alone flags (\cref{supp:attacks}).

\paragraph{Real political footage beyond CivicBench} \method also leads on external political footage, and framing at 1 gray level makes four of the six published detectors flag every real statement. DF26~\cite{shykula2026df26} pairs 71 real official statements with 213 text-to-video fakes from three open generators. Against these fakes, \method reaches an AUC of .715 and the published detectors at most .462. \method flags 14\% of the statements after the same attack and 10\% before it (\cref{supp:genvideo}).

\subsection{Ablations}
\label{sec:ablations}

\begin{table}[t]
\centering
\caption{\textbf{Ablations on CivicBench.} Clean AUC and TPR on all generated clips that each variant scores. Evasion and framing on the first 120 attacked clips of each class, so the \method row differs from \cref{tab:main}. The probe is Mavali et al.\ on \dclip. Best in bold. $^{\dagger}$Not significantly worse than \method ($p\geq.05$, paired, uncorrected). More variants are in \cref{tab:ablation_civic_more}.}
\label{tab:ablation}
\footnotesize
\setlength{\tabcolsep}{2.4pt}
\begin{tabular}{l cc cc cc}
\toprule
& \multicolumn{2}{c}{clean} & \multicolumn{2}{c}{evasion $\uparrow$} & \multicolumn{2}{c}{framing $\downarrow$}\\
\cmidrule(lr){2-3}\cmidrule(lr){4-5}\cmidrule(lr){6-7}
Variant & AUC & TPR & 2 & 4 & 2 & 4\\
\midrule
\method & .941 & .779 & .692 & .593 & .138 & .211\\
w/o margin term & .938 & .757 & .673$^{\dagger}$ & .546 & .165$^{\dagger}$ & .257\\
Madry & .935 & .747 & .691$^{\dagger}$ & .551 & .147$^{\dagger}$ & .239$^{\dagger}$\\
TRADES & .916 & .715 & .634 & .540 & .174 & .284\\
MART & .905 & .685 & .598 & .492 & .165$^{\dagger}$ & .229$^{\dagger}$\\
\midrule
w/o weight averaging & .929 & .751 & .671$^{\dagger}$ & \textbf{.610}$^{\dagger}$ & .138$^{\dagger}$ & .284\\
fakes attacked only & \textbf{.946} & \textbf{.785} & \textbf{.714}$^{\dagger}$ & .559 & .156$^{\dagger}$ & .266\\
w/o fine-tuning (probe) & .638 & .182 & .134 & .108 & \textbf{.046}$^{\dagger}$ & \textbf{.101}$^{\dagger}$\\
\bottomrule
\end{tabular}
\end{table}

Among the parts of \method, fine-tuning matters most. \Cref{tab:ablation} changes one part at a time on CivicBench. The probe on the same frozen encoder detects .182 of the clean fakes against .779. On the 27 attacked fakes that both flag, \method also keeps 37.0 percentage points more at 4 gray levels ($p=.001$). The probe is framed least, consistent with its small reach (\cref{supp:theory}). With weight averaging, standard adversarial objectives come close on clean clips. Under attack, none of them beats \method in any setting, and each falls behind it in at least one. Attacking only fakes and dropping weight averaging mainly raise framing at 4 gray levels (.266 and .284, against .211; \cref{supp:ablation_aigvd}).

\paragraph{What the margin term does} The margin term protects the fakes nearest the threshold, not the typical clip. On the 108 attacked fakes that both models flag, removing it costs 9.3 percentage points at 4 gray levels ($p=.010$). On 109 real clips, removing it adds 4.6 points of framing ($p=.011$, uncorrected). In three training runs of each recipe, every run of \method keeps more attacked fakes at 4 gray levels than every run without the term or with Madry's objective, and flags fewer attacked real clips than every run without the term (\cref{supp:attacks}). By \cref{prop:survival} the gain should lie in the fakes nearest the threshold. On the same 108 fakes, the 10th percentile of the break budget is 3.0 gray levels with the margin term and 1.9 without. The 95\% CI of the difference is [+0.3, +2.1]. The medians are similar (7.7 and 7.6, \cref{fig:reach}c).

\section{Conclusion}
\label{sec:conclusion}

Detectors of political video face two attacks, hiding fakes and framing real footage. CivicBench measures both. At 1 gray level, published detectors keep at most .045 of the fakes flagged, and four of them flag nearly every attacked real clip. Tests of evasion alone miss this second failure. \method keeps the reach small with a robust encoder and adds a margin at its deployed threshold. At 2 gray levels, it keeps .668 of the fakes flagged, against at most .289 for published detectors, and framing raises its false alarms only from 8\% to 14\%. Margin against reach also explains why detectors that read codec motion vectors fail. We therefore suggest testing detectors in both directions, at their deployed threshold. We release CivicBench and its toolkit so that this test takes one command.

\begin{availability}
CivicBench is hosted on Hugging Face at \url{\HFURL}. The code and the attack toolkit are at
\url{https://github.com/KurbanIntelligenceLab/CivicBench}. The toolkit installs with \texttt{pip install civicbench}.
\end{availability}
\begin{llmusage}
Large language models (Claude, Anthropic) assisted with writing and editing, finding and verifying related work and references, drafting the propositions and proofs, and proposing experiments. Every reported number was computed from experiment records by analysis code, not generated by a language model. The authors set the research direction, reviewed all AI-assisted content, and tested the formal claims against measurements. They take full responsibility for the paper and any remaining errors.
\end{llmusage}

\bibliography{references}

\appendix
\noindent This appendix follows the order of the paper. \Cref{supp:civic} describes CivicBench in detail, with its prompts, sources, controls, results per generator, scene and level, and a datasheet with the release plan. \Cref{supp:proofs} proves the propositions of \cref{sec:theory,sec:method}. \Cref{supp:theory} measures margin against reach on CivicBench and on the general benchmark \aigvd. By \cref{prop:survival}, a clip keeps its label under attack exactly when its margin from the threshold exceeds the reach, how far the attack can move its score. \Cref{supp:mvties} gives the measurements behind the failure of motion-vector detectors. \Cref{supp:setup} gives the full setup, the training step, and the attacks. \Cref{supp:ci} gives confidence intervals, paired tests, and five-fold results. \Cref{supp:attacks} covers stronger attacks, uploads, perceptual cost, seeds, and the detectors of an audit toolkit. \Cref{supp:genvideo} covers the external political footage and the other unseen test sets. \Cref{supp:ablation_aigvd} gives further ablations. \Cref{supp:compute} covers compute and broader impact.

\section{CivicBench in Detail}
\label{supp:civic}

\paragraph{Terms} A detector flags a clip as generated when its score reaches a threshold $\tau$. AUC is the area under the receiver operating characteristic (ROC) curve, the chance that a random generated clip scores above a random real one. TPR, the true-positive rate, is the share of generated clips flagged at a threshold set for 5\% false alarms on real clips. A gray level, the unit of the attack budget, is one step on the 0 to 255 scale of a pixel. Evasion pushes the score of a generated clip below the threshold, and framing pushes the score of a real clip above it. PGD is projected gradient descent, which uses the detector's gradients. The Square attack needs only the detector's scores.

\paragraph{Detectors and training recipes} Unless stated otherwise, models are trained on fold 0 of \aigvd (\cref{sec:setup}). The detectors are those of \cref{tab:main}, described in \cref{sec:setup}. The CLIP probe is a linear probe on the CLIP encoder ViT-L/14, a vision transformer (ViT). R2 and R4 are ViT-L/14 CLIP encoders that Schlarmann et al.~\cite{schlarmann2024robust} fine-tuned adversarially at 2 and 4 gray levels. \dclip is the ViT-H/14 encoder that Wang et al.~\cite{wang2025double} pretrained on attacked images. In the tables, ``Mavali, R2'' is the detector of Mavali et al.\ on R2, and likewise for R4 and \dclip. Madry's objective~\cite{madry2018towards}, TRADES~\cite{zhang2019theoretically}, and MART~\cite{wang2020improving} are standard losses of adversarial training. In our comparisons they replace \method's loss. They keep its encoder, fine-tuning, training attack, and weight averaging (\cref{sec:method,supp:setup}). The weight averaging is an exponential moving average of the weights. TRADES and MART also drop \method's generator weights. ``Madry'' in the tables is Madry's objective trained this way.

\begin{table}[t]
\centering
\caption{\textbf{Real footage of CivicBench.} Videos, clips, scene kinds, video licenses, and range of upload years per source. PD is public domain and CC Creative Commons. Other is the National Assembly of Venezuela and TV Senado of Brazil.}
\label{tab:civic_sources}
\scriptsize
\setlength{\tabcolsep}{3pt}
\begin{tabular}{l rr l l c}
\toprule
Source & Videos & Clips & Scene kinds & Licenses (videos) & Years\\
\midrule
Demonstrations (Wikimedia Commons) & 419 & 838 & protest, speech & CC BY-SA 271, CC BY 125, CC0 16, PD 7 & 2011--2026\\
White House (Commons) & 367 & 720 & briefing, speech & PD 366, CC BY-SA 1 & 2011--2026\\
US State Department (Internet Archive) & 294 & 588 & briefing & PD 294 & 2022--2025\\
US Senate (Commons) & 155 & 309 & legislature & PD 155 & 2019--2026\\
Kremlin.ru (Commons) & 75 & 150 & speech & CC BY 74, CC BY-SA 1 & 2000--2025\\
Other (Commons) & 14 & 27 & legislature & PD 11, CC BY 3 & 2016--2026\\
\midrule
all & 1,324 & 2,632 & & & \\
\bottomrule
\end{tabular}
\end{table}

\begin{figure}[t]
\centering
\includegraphics[width=\linewidth]{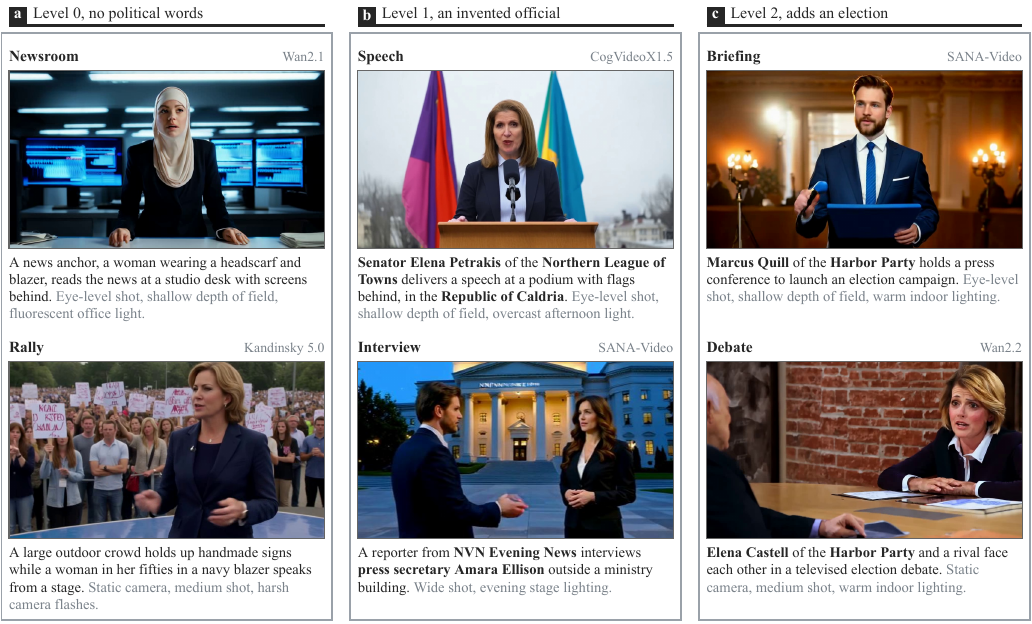}
\caption{\textbf{Example prompts of CivicBench}, two per level, each with the middle frame of one generator's clip. Bold words are filled in from the lists of names, roles, parties, countries, and channels. Gray words give the camera and lighting. Every prompt also contains ``Realistic news footage'', left out here.}
\label{fig:prompts}
\end{figure}

\paragraph{Prompts} Templates describe nine kinds of civic scenes, namely speeches, press briefings, legislatures, rallies, protests, newsrooms, interviews, debates, and crisis reports. Each kind has templates at three levels of political content. Level 0 shows the scene with no political words, level 1 names an invented official and role, and level 2 adds an election. The templates are filled from fixed lists of 24 first and 19 last names, 10 roles, and 8 parties. Further lists hold 6 countries, 4 news channels, 6 camera descriptions, and 6 lighting descriptions. The full lists are part of the release. The templates produced 2,700 prompts, 300 per scene kind. Each prompt was assigned level 0, 1, or 2 at random. The probabilities were 35\%, 45\%, and 20\%, so that level 1, the generic political case, is the most common. Duplicate prompts were redrawn. Level-0 templates have no names to vary, so they produce the most duplicates. As a result, 25\% of the 2,700 prompts are at level 0, 51\% at level 1, and 23\% at level 2. The slowest generators take 2 to 14 minutes a clip. To halve the cost of generation, CivicBench uses every second prompt, in order of prompt id, within each kind and level. This leaves 1,356 prompts, 151 or 150 per kind. Of these, 344 are at level 0, 693 at level 1, and 319 at level 2. The identity screen (\cref{supp:identity}) later removes 6 of them, so 1,350 prompts enter the benchmark. Wan2.2, Wan2.1, and LTX-Video also rendered the other half, which we keep as a replication set. Every prompt ends with ``Realistic news footage'' and a camera and lighting description, to ask for the look of news video. Each prompt has a fixed seed for reproducibility. The generators differ in architecture, so a shared seed does not make their clips alike. Comparisons across generators rest on the shared prompt instead. A template body is a prompt without its camera and lighting description. The 1,356 prompts hold 964 distinct bodies. Some combinations of scene kind and level (crisis level 0, debate level 0, protest level 0) have a single body each, varied only by camera and lighting. Their 54 prompts are near-duplicates. We therefore compute confidence intervals (CIs) per scene kind and level by resampling template bodies, not prompts. For example, \method's TPR at level 0 is .772 with 95\% CI [.722, .815]. \Cref{fig:prompts} shows two prompts per level, each with a frame that one of the generators rendered from it.

\paragraph{Why these scenes} The nine scene kinds are meant to cover the ways in which a political deepfake poses as footage of a real event. We chose them before the check below, which confirms rather than selects them. We read the 3,053 reports of the Political Deepfakes Incidents Database (PDID)~\cite{walker2024merging} and kept the 1,059 videos that carry a summary and are coded as deepfakes or as uncertain. Keyword rules on the title and summary assigned each video a scene kind. We checked 60 random videos by hand, and the rules agree with our reading on 78\% of them. Once we correct for the errors of the rules, about 39\% of the videos show one of the nine kinds. Speeches alone make up 25\% of all videos. The rest are mostly satire or fantasy, such as politicians dancing or cast in films. Such videos do not pose as footage, and a detector need not flag them to protect the public record. One recurring kind falls outside the nine. It shows staged conversations of ordinary voters and is closest to our interviews. CivicBench has real footage for four of the kinds, namely speeches, briefings, legislatures, and protests. They make up 74\% of the videos that show one of the nine kinds. Governments, legislatures, and Wikimedia Commons also publish these kinds under public-domain or open licenses. Broadcast news, interviews, and debates are mostly copyrighted. Uploads of Voice of America are public domain but embed footage from news agencies. We also weight each kind by its share of the coded PDID videos. This weighting raises the TPR of most detectors by up to 8 percentage points. It does not change the order of the detectors listed next. With this weighting, \method detects .840 of the generated clips and Madry's objective .833. The CLIP probe detects .412 and \stream .309.

\paragraph{Names} Some of our invented names belong to real people. We searched Wikidata for every name that the templates can form, by exact match of its English label. Of the 456 pairs of first and last names, 22 are also names of real people, mostly private individuals or historical figures. Among them, 3 are names of people who held a public position. In version 1.0 of CivicBench, names of real people appeared in 43 of the 1,356 prompts. Names of people who held a public position appeared in 7. Two party names were also real. The Unity Party is the name of parties in more than 10 countries and appeared in 44 prompts. The Civic Alliance is a party in Bosnia and Herzegovina and appeared in 63 prompts. Version 1.1, which this paper uses, replaces the three names of officeholders and the two party names with invented names that have no Wikidata match. The new names are Daniel Ostervald, Daniel Quennell, Grace Halvorine, the Meadowmark Party, and the Larkspur Alliance. The change touches 113 of the 1,356 prompts. We regenerated their 904 clips with the original settings and seeds and scored them again. In each attack set, a regenerated clip stays and is attacked again if the detector whose flagged clips form the set still flags it. Otherwise, another flagged clip of the same generator takes its place. No country or channel name is that of a real country or channel. Ostrava Union, however, borrows the name of a Czech city, and the Northern League of Towns recalls Italy's Lega Nord. A name only shapes the prompt. The identity screen below checks the generated faces themselves.

\begin{table}[t]
\centering
\caption{\textbf{CivicBench controls.} Left, clean AUC, 95\% CI, and TPR of models trained on \aigvd. Middle, their AUC when the real clips are only those sampled at exactly 8 fps, or 924 real clips re-encoded to 480p. Right, models trained on \gvb. D3 needs no training, so both sides show the same model.}
\label{tab:civic_src}
\footnotesize
\setlength{\tabcolsep}{4pt}
\begin{tabular}{l ccc cc cc}
\toprule
& \multicolumn{3}{c}{trained on \aigvd} & \multicolumn{2}{c}{controls (AUC)} & \multicolumn{2}{c}{trained on \gvb}\\
\cmidrule(lr){2-4}\cmidrule(lr){5-6}\cmidrule(lr){7-8}
Detector & AUC & 95\% CI & TPR & 8 fps & 480p & AUC & TPR\\
\midrule
CLIP probe & .781 & [.770, .793] & .364 & .796 & .748 & .787 & .381\\
\restrav & .448 & [.430, .466] & .009 & .679 & .433 & .684 & .075\\
D3 & .475 & [.457, .493] & .028 & .707 & .457 & .475 & .028\\
\stream & .714 & [.702, .726] & .309 & .759 & .679 & .805 & .352\\
\streammv & .602 & [.590, .616] & .111 & .675 & .573 & .734 & .214\\
\mvmob & .587 & [.573, .601] & .086 & .666 & .548 & .717 & .221\\
\midrule
Mavali, R2 & .768 & [.756, .780] & .381 & .802 & .722 & .742 & .323\\
Mavali, R4 & .774 & [.762, .786] & .394 & .812 & .731 & .678 & .244\\
Mavali, \dclip & .638 & [.624, .655] & .182 & .805 & .662 & .708 & .239\\
Shahjahan et al. & .709 & [.696, .722] & .233 & .611 & .680 & .598 & .113\\
\midrule
\method & .941 & [.936, .946] & .779 & .957 & .944 & .791 & .299\\
\bottomrule
\end{tabular}
\end{table}

\paragraph{Generator settings} \Cref{tab:civic_gen} lists what each model rendered. To fit the 10,848 rendered clips into our compute, seven of the eight models render shorter clips than their documented defaults. Wan2.2 renders 49 frames instead of 121, and Kandinsky 5.0 renders 61 instead of 121. Wan2.1 and CogVideoX1.5 render 33 instead of 81, and LTX-Video renders 49 instead of its default of 161. Each clip is long enough to supply the 11 frames that detectors read, at about 8 frames per second (fps). Most models also use fewer denoising steps than their pipeline default. Wan, LTX-Video, and CogVideoX1.5 use 40 steps instead of 50. Mochi uses 40 instead of 64, and HunyuanVideo-1.5 uses 30. Guidance scales range from 3 to 6. Two settings did not take effect. HunyuanVideo-1.5 used its default guidance scale instead of ours, and SANA-Video rendered its default of 81 frames instead of our frame count. Fewer steps could leave more artifacts and make clean detection easier. We therefore rendered 87 prompts of each of Wan2.2, CogVideoX1.5, and LTX-Video again with the model's default steps and length. Across detectors and these generators, the TPR at the CivicBench threshold changes by $-$.080 to +.126, with no consistent direction. \method's TPR changes by $-$.069 to +.092. A ninth model, Allegro, was piloted and left out because its decoded clips were color noise.

\paragraph{Real footage} We listed candidate videos from Wikimedia Commons categories and the Internet Archive in October 2026. We kept a video when it was listed at least 1,280 pixels wide (720p on the Internet Archive), at least 20 seconds long, and under an open license. Uploads of Voice of America were dropped because they mix in footage from news agencies. Commons clips are cut from each video's 720p rendition, so no clip is enlarged to the 512 pixels that detectors read. We set quotas by hand at 450 speech, 300 briefing, 175 legislature, and 425 protest videos. At most 75 videos come from the Kremlin. At two clips per video, these quotas give 2,700 clips, the same number as the prompts. The legislature quota is the smallest because only short Commons files could be fetched. The two clips of a video start at evenly spaced points inside its middle 80\%. Of 2,700 planned clips, 2,672 were cut. The other 28 could not be downloaded. We checked the cut clips by motion, frame hashes, and eye. The check found 129 title cards with no footage and the two clips of a video uploaded twice. We replaced each of these 131 clips with a later window of the same video that shows footage (120 clips). When a video had no such window, we used a clip of another video of the same kind (11 clips from 6 videos). A final audit before release removed 40 more clips without replacement. Fourteen show AI-animated portraits that the White House uploaded, 13 show title cards or static graphics, and 13 repeat footage that was uploaded twice. The remaining 2,632 clips come from 1,324 videos. \Cref{tab:civic_sources} lists the sources with their licenses and upload years. The public-domain videos include US federal works. Upload years are parsed from the video titles and are unknown for 665 clips.

\begin{table}[t]
\centering
\caption{\textbf{Frame rate within one source.} AUC of generated clips against one source's real clips, sampled at exactly 8 frames per second (8) or not (other). Frame rate used as the only score reaches an AUC of .942 (White House) and .772 (demonstrations).}
\label{tab:fps_within}
\footnotesize
\setlength{\tabcolsep}{4pt}
\begin{tabular}{l cc cc}
\toprule
& \multicolumn{2}{c}{White House} & \multicolumn{2}{c}{demonstrations}\\
\cmidrule(lr){2-3}\cmidrule(lr){4-5}
Detector & 8 & other & 8 & other\\
\midrule
CLIP probe & .768 & .871 & .818 & .838\\
\restrav & .594 & .298 & .739 & .717\\
D3 & .622 & .346 & .769 & .750\\
\stream & .677 & .733 & .816 & .754\\
\streammv & .634 & .591 & .705 & .658\\
\mvmob & .637 & .612 & .690 & .669\\
Mavali, R2 & .753 & .798 & .837 & .868\\
Mavali, R4 & .756 & .802 & .853 & .866\\
Mavali, \dclip & .715 & .683 & .870 & .873\\
Shahjahan et al. & .716 & .789 & .541 & .656\\
Madry & .939 & .964 & .947 & .934\\
\method & .948 & .971 & .963 & .947\\
\bottomrule
\end{tabular}
\end{table}

\begin{table}[t]
\centering
\caption{\textbf{False alarms without attack.} Share of real clips flagged, per source. Deployed rows use the threshold that each detector was trained with on \aigvd, and CivicBench rows use CivicBench's own 5\% threshold. Small type in the all column gives the 95\% CI over source videos.}
\label{tab:fpr_source}
\footnotesize
\setlength{\tabcolsep}{3pt}
\begin{tabular}{l l ccccccc}
\toprule
Threshold & Detector & all & White House & State Dept. & US Senate & Kremlin & protests & other\\
\midrule
deployed & CLIP probe & 10.6\% {\tiny[9.2, 12.1]} & 6.6\% & 23.4\% & 13.3\% & 10.7\% & 4.1\% & 13.2\%\\
deployed & \restrav & 15.1\% {\tiny[13.7, 16.7]} & 18.9\% & 25.9\% & 17.5\% & 16.0\% & 3.1\% & 18.4\%\\
deployed & \stream & 14.0\% {\tiny[12.6, 15.6]} & 13.2\% & 19.5\% & 14.6\% & 13.3\% & 10.9\% & 13.2\%\\
deployed & \streammv & 11.4\% {\tiny[10.1, 12.6]} & 12.5\% & 15.9\% & 7.5\% & 18.0\% & 7.9\% & 2.6\%\\
deployed & \mvmob & 12.0\% {\tiny[10.6, 13.5]} & 8.3\% & 20.6\% & 11.4\% & 16.0\% & 8.7\% & 13.2\%\\
deployed & Mavali, R2 & 13.8\% {\tiny[12.2, 15.4]} & 12.6\% & 24.7\% & 17.5\% & 18.7\% & 4.5\% & 21.1\%\\
deployed & Mavali, R4 & 10.7\% {\tiny[9.2, 12.1]} & 9.4\% & 17.4\% & 17.5\% & 14.0\% & 3.3\% & 23.7\%\\
deployed & Mavali, \dclip & 35.1\% {\tiny[32.6, 37.5]} & 19.2\% & 91.5\% & 45.1\% & 56.7\% & 1.9\% & 28.9\%\\
deployed & Shahjahan et al. & 37.0\% {\tiny[35.0, 39.2]} & 27.5\% & 22.2\% & 45.1\% & 38.0\% & 52.9\% & 28.9\%\\
deployed & Madry & 6.6\% {\tiny[5.5, 7.7]} & 5.3\% & 2.4\% & 9.1\% & 22.7\% & 6.3\% & 15.8\%\\
deployed & \method & 6.4\% {\tiny[5.4, 7.6]} & 3.6\% & 7.2\% & 11.4\% & 12.0\% & 4.7\% & 23.7\%\\
\midrule
CivicBench & Madry & 5.0\% {\tiny[4.1, 6.0]} & 3.5\% & 1.9\% & 6.5\% & 16.7\% & 5.5\% & 13.2\%\\
CivicBench & \method & 5.0\% {\tiny[4.1, 6.0]} & 2.9\% & 5.1\% & 9.7\% & 7.3\% & 3.7\% & 23.7\%\\
\bottomrule
\end{tabular}
\end{table}

\paragraph{Shortcuts} Before processing, clip length alone separates real and generated clips almost perfectly. Real clips run 4 seconds, and generated ones 1.43 to 5.06 seconds. Detectors never see the length, because every clip is reduced to 11 frames. The release therefore makes the 11-frame inputs the official test input. The sampled frame rate, however, remains a cue. Generated clips are sampled at 8 fps, Mochi's at 7.5. Real ones are sampled at 6.0 to 10.0 fps, depending on the frame rate of their source. Of these, 224 are sampled at 8.56 fps, from 59.94-fps sources. The rate alone separates real and generated clips with an AUC of .895. Almost all of the 518 real clips sampled at exactly 8 fps come from the White House and from demonstrations. For the 8 fps column of \cref{tab:civic_src}, we score the generated clips against these clips alone, so this comparison changes the mix of sources as well as the frame rate. In \cref{tab:fps_within}, we therefore hold the source fixed. Within one source, we score the generated clips against the real clips sampled at exactly 8 fps and, separately, against the other real clips. Only the White House and the demonstrations have more than 50 clips at that rate. The White House has 207 such clips and 507 others. The demonstrations have 306 and 530. The other sources have at most 4. The two groups of demonstration clips differ mainly in frame rate. Between them, no detector's AUC changes by more than .12, and \method's changes by .02. Among White House clips, the rate alone reaches an AUC of .942. There, \restrav and D3 separate the generated clips from the real clips at 8 fps far better than from the other real clips, by up to .30 in AUC. Both read change between frames. They score the White House clips sampled at 7.5 fps like generated clips. Part of their below-chance AUC on CivicBench thus comes from frame rate, not from what they detect.

\paragraph{Processing control} The processing history of the real clips does not explain the low clean AUC of the published detectors. The real clips come from videos at least 1,280 pixels wide that were compressed before we cut them. The generators, in contrast, write each clip once, at 480 to 768 pixels high. To give real clips a generator-like history, we re-encoded 924 of them once with H.264 at 480 pixels high, as a generator writes its output. We then passed them through the same processing. Against these real clips, no detector's AUC changes by more than .05, and \method's is .944 (\cref{tab:civic_src}). The constant rate factor (CRF) is the H.264 quality setting. This control encodes at a CRF of 23, while the generators wrote their clips at about CRF 25. We also encoded all 2,501 real clips at CRF 25 instead. This changes no detector's AUC by more than .05, and \method's by .002.

\paragraph{Training benchmark} \method's lead over published detectors holds when every detector uses its best training set, but its rank among the adversarial recipes depends on the training set. \Cref{tab:civic_src} gives the results of the models trained on fold 0 of \aigvd, as in \cref{tab:main}, with 95\% CIs from a bootstrap over prompts and source videos. \gvb is the other general benchmark (\cref{sec:setup}). Detectors trained on fold 0 of \gvb transfer differently to CivicBench. The four generators of \gvb date from 2023. Trained on \gvb, the detectors that read motion and the encoded stream do better than when trained on \aigvd, and so do \restrav and Mavali et al.\ on \dclip. The CLIP probe changes little. \method and Mavali et al.\ on R2 and R4 do worse. Trained on \gvb, \method reaches an AUC of .791, on par with \stream. Trained on both benchmarks, it reaches .941. When each detector uses the training set that suits it best, no published detector exceeds .805 (\stream trained on \gvb). Trained on \gvb, however, three recipes reach a higher AUC than \method. Two of them are TRADES with \method's margin term (.847, \cref{eq:margin}) and \method without weight averaging (.832). The third is Madry's objective without weight averaging (.806). Trained on both benchmarks, TRADES with the margin term reaches .949. The main paper reports the models trained on \aigvd, because its generators are closer to those of CivicBench. A deployed detector would likewise be trained on generators close to those that an attacker uses.

\begin{table}[t]
\centering
\caption{\textbf{One real source left out.} AUC on CivicBench with all real clips (all) and with the real clips of one source left out. WH is the White House, State the State Department, and Senate the US Senate.}
\label{tab:civic_loso}
\scriptsize
\setlength{\tabcolsep}{2.6pt}
\begin{tabular}{l ccccccc}
\toprule
Detector & all & WH & State & Senate & Kremlin & protests & other\\
\midrule
CLIP probe & .781 & .759 & .818 & .783 & .784 & .758 & .781\\
\restrav & .448 & .472 & .509 & .465 & .457 & .318 & .449\\
D3 & .475 & .493 & .528 & .502 & .487 & .343 & .475\\
\stream & .714 & .712 & .744 & .713 & .713 & .684 & .714\\
\streammv & .602 & .602 & .627 & .601 & .608 & .569 & .602\\
\mvmob & .587 & .575 & .624 & .593 & .589 & .545 & .587\\
Mavali, R2 & .768 & .762 & .797 & .773 & .774 & .727 & .769\\
Mavali, R4 & .774 & .768 & .795 & .783 & .779 & .733 & .775\\
Mavali, \dclip & .638 & .619 & .731 & .647 & .648 & .530 & .639\\
Shahjahan et al. & .709 & .688 & .690 & .712 & .708 & .753 & .708\\
Madry & .935 & .927 & .939 & .935 & .939 & .933 & .935\\
\method & .941 & .933 & .949 & .943 & .943 & .936 & .941\\
\bottomrule
\end{tabular}
\end{table}

\begin{table}[t]
\centering
\caption{\textbf{AUC per generator on CivicBench.} Each generator's clips are scored against all real clips. Models are trained on fold 0 of \aigvd.}
\label{tab:civic_pergen}
\scriptsize
\setlength{\tabcolsep}{4pt}
\begin{tabular}{l cccccccc}
\toprule
Detector & Wan2.2 & Wan2.1 & LTX-Video & Mochi & CogVideoX1.5 & HunyuanVideo-1.5 & SANA-Video & Kandinsky 5.0\\
\midrule
CLIP probe & .716 & .785 & .753 & .854 & .769 & .664 & .966 & .744\\
\restrav & .494 & .512 & .298 & .426 & .443 & .418 & .583 & .410\\
D3 & .520 & .522 & .194 & .423 & .465 & .479 & .655 & .541\\
\stream & .591 & .828 & .713 & .873 & .514 & .634 & .948 & .610\\
\streammv & .578 & .629 & .732 & .706 & .548 & .529 & .600 & .497\\
\mvmob & .557 & .707 & .699 & .555 & .593 & .471 & .683 & .428\\
Mavali, R2 & .683 & .773 & .622 & .780 & .712 & .752 & .969 & .856\\
Mavali, R4 & .711 & .804 & .588 & .793 & .691 & .768 & .971 & .864\\
Mavali, \dclip & .618 & .701 & .332 & .670 & .631 & .630 & .881 & .645\\
Shahjahan et al. & .683 & .721 & .544 & .830 & .635 & .703 & .879 & .678\\
Madry & .875 & .945 & .920 & .929 & .938 & .912 & .996 & .965\\
\method & .880 & .944 & .915 & .930 & .948 & .933 & .999 & .981\\
\bottomrule
\end{tabular}
\end{table}

\paragraph{One source left out} No clean comparison in the paper depends on a single real source. Each kind of real scene comes mostly from one source. Briefings come from the State Department, legislatures from the US Senate, speeches from the White House, and protests from demonstrations on Wikimedia Commons. Results per scene kind are therefore also results per source. In \cref{tab:civic_loso}, we leave out the real clips of one source at a time. The TPRs below use a threshold reset on the remaining real clips. The AUC of \method stays between .933 and .949. That of the CLIP probe stays between .758 and .818, and that of \stream between .684 and .744. The State Department footage affects the published detectors most. Without it, \stream's AUC rises from .714 to .744, while its TPR stays at .310. The CLIP probe's TPR rises from .364 to .422, while \method's stays at .782. Most US Senate clips carry the logo of C-SPAN, and many generated news scenes show invented captions. Without the US Senate clips, \method's AUC is .943.

\paragraph{False alarms per source} At their deployed thresholds, the published detectors and those built to resist attacks flag more real clips than \method. For each source, \cref{tab:fpr_source} gives the share of real clips that each detector flags without any attack. It uses two thresholds. The deployed threshold is the one that each detector was trained with on \aigvd. CivicBench's own threshold is set for 5\% false alarms on its real clips. At the deployed threshold, the published detectors and those built to resist attacks flag 10.6\% to 37.0\% of all real clips, and \method 6.4\%. In the table, small type gives the 95\% CI for all sources together, from a bootstrap over source videos. The White House supplies 720 clips and the State Department 588. The US Senate supplies 309 and the Kremlin 150. Demonstrations supply 838 clips (column protests), including 32 clips of conference talks that we label as speeches. The other 27 come from the National Assembly of Venezuela and TV Senado of Brazil.

\paragraph{Results per generator, scene, and level} \Cref{tab:civic_pergen} gives every detector's AUC per generator. \Cref{tab:civic_scene} gives each detector's TPR per scene kind and per level of political content, at the 5\% threshold.

\begin{figure}[t]
\centering
\includegraphics[width=\linewidth]{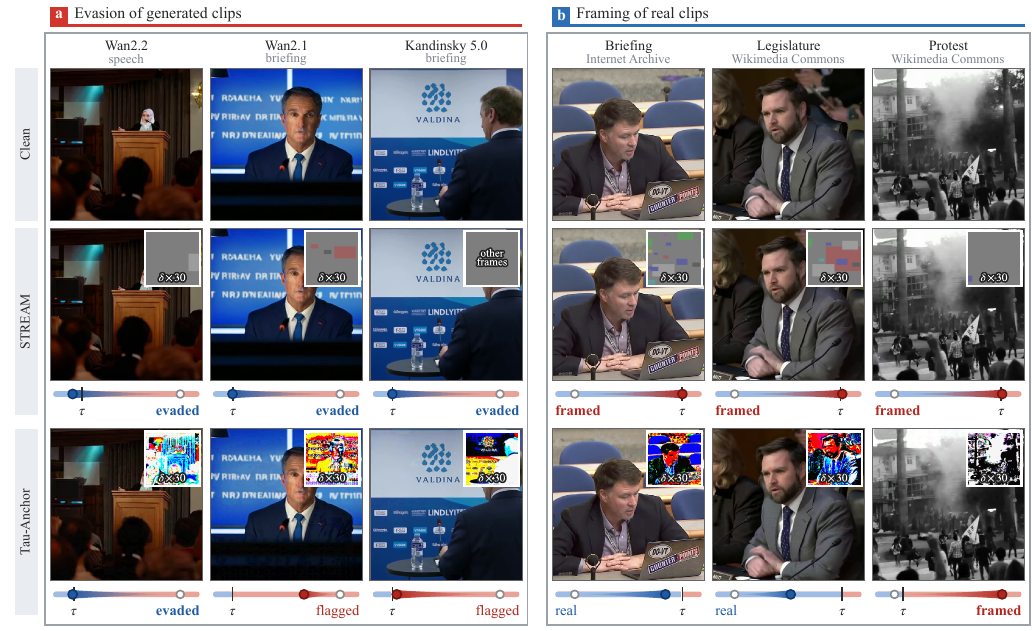}
\caption{\textbf{Attacked CivicBench clips.} (a)~Generated clips under evasion. (b)~Real clips under framing. Rows show the middle frame of each clip, clean and after the attacks on \stream (Square, 1 gray level) and \method (PGD, 4 gray levels). Insets show each attack's change, magnified 30 times. Gauges show the clean score (open), the attacked score (filled), and the threshold $\tau$. Frames keep their aspect ratio and are cropped to a square. The detectors see them squeezed to $512\times512$ pixels.}
\label{fig:examples_civic}
\end{figure}

\paragraph{Attacked examples} One gray level breaks \stream on all six clips of \cref{fig:examples_civic}, while 4 gray levels break \method on two of them. The figure shows three generated clips under evasion and three real clips under framing. Each clip is attacked against \stream with the Square attack at 1 gray level and against \method with PGD at 4 gray levels. For each clip, the figure shows the middle frame, clean and after each attack. The inset in the corner of an attacked frame shows the change that the attack made to this frame, magnified 30 times. The label ``Other frames'' marks a Square attack that left the middle frame unchanged. \method keeps two of the three generated clips flagged and two of the three real clips unflagged. At 4 gray levels, PGD evades it on the generated speech clip and frames the protest clip.

\begin{table}[t]
\centering
\caption{\textbf{TPR per scene kind and level.} Detectors at their 5\% threshold on CivicBench. Real footage covers only the first four scene kinds. Levels of political content are 0 (no political words), 1 (an invented official), and 2 (an election).}
\label{tab:civic_scene}
\scriptsize
\setlength{\tabcolsep}{3pt}
\begin{tabular}{l ccccccccc ccc}
\toprule
& \multicolumn{9}{c}{scene kind} & \multicolumn{3}{c}{level}\\
\cmidrule(lr){2-10}\cmidrule(lr){11-13}
Detector & speech & briefing & legislature & protest & rally & newsroom & interview & debate & crisis & 0 & 1 & 2\\
\midrule
CLIP probe & .449 & .450 & .495 & .225 & .228 & .501 & .438 & .287 & .205 & .457 & .331 & .339\\
\restrav & .011 & .012 & .020 & .001 & .001 & .023 & .007 & .006 & .002 & .012 & .008 & .009\\
D3 & .024 & .027 & .066 & .018 & .000 & .059 & .007 & .048 & .002 & .031 & .030 & .019\\
\stream & .307 & .283 & .338 & .255 & .248 & .348 & .283 & .433 & .285 & .302 & .310 & .317\\
\streammv & .123 & .138 & .089 & .075 & .146 & .139 & .083 & .129 & .080 & .101 & .110 & .130\\
\mvmob & .061 & .083 & .055 & .083 & .127 & .125 & .044 & .113 & .082 & .073 & .090 & .093\\
Mavali, R2 & .474 & .490 & .595 & .156 & .251 & .485 & .403 & .377 & .201 & .449 & .350 & .380\\
Mavali, R4 & .516 & .523 & .568 & .142 & .216 & .496 & .410 & .422 & .249 & .472 & .360 & .388\\
Mavali, \dclip & .500 & .321 & .278 & .013 & .025 & .130 & .077 & .255 & .037 & .181 & .207 & .128\\
Shahjahan et al. & .329 & .347 & .256 & .173 & .197 & .236 & .231 & .228 & .099 & .246 & .238 & .212\\
Madry & .884 & .843 & .896 & .479 & .575 & .885 & .703 & .867 & .588 & .778 & .723 & .782\\
\method & .885 & .877 & .907 & .584 & .641 & .910 & .753 & .859 & .595 & .772 & .763 & .826\\
\bottomrule
\end{tabular}
\end{table}

\paragraph{Datasheet} We answer the questions of Gebru et al.~\cite{gebru2021datasheets} in brief. \emph{Motivation.} CivicBench tests detectors of generated video on political footage, clean and under attacks in both directions. The authors built it for this paper. \emph{Composition.} CivicBench holds 10,800 generated clips of invented politicians from eight open generators (\cref{tab:civic_gen}) and 2,632 real clips from 1,324 videos (\cref{tab:civic_sources}). Each generated clip carries its generator, prompt, scene kind, and level. Each real clip carries its source, license, scene kind, and, where known, upload year. The real clips show officials and members of the public at public events. The generated clips are designed to show no real person. The identity screen below found 23 that do resemble one. We replace 17 of them with clean new renders and drop the other 6 prompts for all eight generators. \emph{Collection.} Prompts come from the templates above, and real videos were listed and cut as described above. \emph{Processing.} Every clip is re-encoded with H.264 and sampled to 11 frames of $512\times512$ pixels (\cref{supp:setup}). The unprocessed clips are kept. The 11-frame inputs are the official test input, since raw lengths differ between real and generated clips. \emph{Uses.} CivicBench is for testing detectors of generated video, clean and under attack. It is a test set. It should not be used to train the detectors that it then tests, or to train generators. \emph{Distribution and maintenance.} The authors maintain CivicBench and handle requests for removal, through a contact address given with the release. Version 1.1, used here, replaces five invented names that matched real ones and 131 real clips that showed no footage or repeated another clip. It also removes 40 real clips found in a final audit, as described above. Version 1.0 is archived.

\paragraph{Release} The real clips stay under their own licenses. We host them openly, with no request form. We add no terms, because the licenses of the 475 videos under CC BY or CC BY-SA forbid added terms. Each clip carries its creator, license, source page, start time, and the changes we made. Clips cut from CC BY-SA videos stay under CC BY-SA. The official test input is stored as lossless frames, with checksums. The generated clips are ours to license. We release them openly, for research on detection, with a contact for takedown requests. HunyuanVideo-1.5's license forbids using its outputs to improve other models, so its clips are for testing only.

\paragraph{Toolkit} A Python package, installable with pip under the Apache-2.0 license, loads CivicBench and runs the attacks of this paper on any detector, at any budget. The detector is given as a function from frames to a score. The package reports the columns of \cref{tab:main} for the chosen budgets. Its source is at \url{https://github.com/KurbanIntelligenceLab/CivicBench}, and it installs with \texttt{pip install civicbench}.

\paragraph{Ethics} The real footage shows officials at public events. The 838 demonstration clips also show members of the public. We use only footage that its publishers released for reuse, and we identify no one. We did not seek review by an institutional review board (IRB), because the work involves no interaction with people and collects no private information. For the same reason, we did not seek the consent of the people shown. They appear at public events, and governments or the uploaders recorded the footage and published it under open licenses. Anyone shown can ask for a clip's removal through the release's takedown contact. The release contains no attacked real clips, and thus no ready-made forgeries of real footage. The generated clips show invented politicians, parties, and countries. The names in the prompts come only from fixed lists of invented names. The Wikidata check above reports which of them match real people or parties. No template describes voter suppression, ballot tampering, or violence.

\paragraph{Identity screen}\label{supp:identity} Open generators can reproduce real leaders from prompts that never name them. Names do not control the faces that a generator draws. We therefore detect faces in up to three frames of every generated clip with the YuNet face detector of OpenCV. We compare each face with portraits of 12,900 politicians from Wikidata, using the SFace face recognizer of OpenCV. In all, we compare 41,464 faces from the 8,254 clips in which the detector finds a face. Most close matches are paintings, statues, or dark and blurred faces, and they spread evenly over the eight generators. HunyuanVideo-1.5, however, drew look-alikes of four heads of state and one parliamentary speaker, although no prompt named them. They appear in 22 of its clips and in 1 clip of Mochi. Their best matches reach a cosine similarity of .75, while no two different people in the portrait gallery reach .58. We checked these matches by eye and withhold the 23 clips from the release and from every result. We rendered their prompts again with new seeds and screened each new clip the same way. We then screened every fourth frame of each passing clip and checked its faces by eye. Seventeen prompts gave a clean clip within six seeds. Each new clip takes the place of the withheld one in the release and in every clean result. The attacks use only the original clips. Three prompts gave a head of state at all six seeds, and three gave a speaker who still resembled a politician. We drop these six prompts for all eight generators, which leaves 1,350 clips per generator.

\section{Proofs}
\label{supp:proofs}

\begin{figure}[t]
\centering
\includegraphics[width=0.55\linewidth]{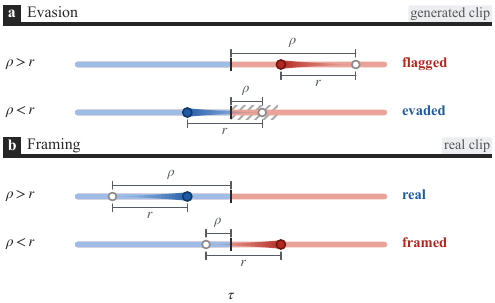}
\caption{\textbf{Margin against reach, schematic.} (a)~Evasion of a generated clip and (b)~framing of a real clip, with margin $\rho$ above the reach $r$ (top row) and below it (bottom row). Gauges show the threshold $\tau$, the clean score (open), the attacked score (filled), and the attack's path. All rows share the same reach. Hatching marks the band of width $\kappa$ above $\tau$ that \method's margin term targets.}
\label{fig:idea}
\end{figure}
This section proves the propositions of \cref{sec:theory,sec:method} in their notation. The set $\B_\eps(x)$ holds the 8-bit clips that differ from $x$ by at most $\eps$ gray levels in every pixel. The margin $\rho$ and the reach $r$ are defined in \cref{sec:margin}, and the gap $g$ in \cref{sec:mv}. \Cref{fig:idea} illustrates \cref{prop:survival}. Its hatched band is where \method's margin term pushes attacked generated clips upward during training (\cref{sec:method}).

\paragraph{\cref{prop:survival}} The proof expresses the score of each reachable clip through the margin and then applies the threshold test to the worst reachable clip. Let $x$ be a flagged generated clip and $x'\in\B_\eps(x)$. Because the head is linear, $s(x')-\tau = s(x)-\tau - w^\top\big(e(x)-e(x')\big) = \norm{w}\big(\rho(x) - w^\top(e(x)-e(x'))/\norm{w}\big)$. So $x'$ is flagged if and only if $w^\top(e(x)-e(x'))/\norm{w}\leq\rho(x)$. The set $\B_\eps(x)$ is finite, because every clip in it has 8-bit pixel values, so the maximum in \cref{eq:disp} exists. The condition therefore holds for every $x'\in\B_\eps(x)$ if and only if it holds for the maximizing clip, that is, if and only if $r(x,\eps)\leq\rho(x)$. Now let $x$ be an unflagged real clip. Then $\tau-s(x') = \norm{w}\big(\rho(x) - w^\top(e(x')-e(x))/\norm{w}\big)$. So $x'$ stays unflagged, $s(x')<\tau$, if and only if $w^\top(e(x')-e(x))/\norm{w}<\rho(x)$. This holds for every $x'\in\B_\eps(x)$ if and only if $r(x,\eps)<\rho(x)$. The rates in the proposition range over all generated or all real clips, including those misclassified before the attack. Since $x$ itself lies in $\B_\eps(x)$, $r(x,\eps)\geq0$. A generated clip that is not flagged therefore has $\rho(x)<0\leq r(x,\eps)$. It fails the condition $\rho(x)\geq r(x,\eps)$ and does not count toward the TPR under evasion. A real clip that is flagged before the attack has $\rho(x)\leq0\leq r(x,\eps)$. It meets the condition $\rho(x)\leq r(x,\eps)$ and counts as framed. Over all clips, a generated clip thus stays flagged for every clip in $\B_\eps(x)$ if and only if $\rho(x)\geq r(x,\eps)$, and a real clip is flagged for some clip in $\B_\eps(x)$ if and only if $\rho(x)\leq r(x,\eps)$. The TPR under evasion and the attacked false-alarm rate are the probabilities of these two events over generated and real clips.

The argument needs no linear head if margin and reach are measured in score units, as differences of scores. Let $s$ be any detector. A flagged generated clip stays flagged on all of $\B_\eps(x)$ if and only if $\min_{x'\in\B_\eps(x)}s(x')\geq\tau$. Equivalently, its margin in score units, $s(x)-\tau$, is at least its reach in score units, $\max_{x'\in\B_\eps(x)}\big(s(x)-s(x')\big)$. Likewise, an unflagged real clip stays unflagged if and only if $\tau-s(x)>\max_{x'\in\B_\eps(x)}\big(s(x')-s(x)\big)$, that is, if its margin in score units exceeds its reach in score units. \Cref{supp:theory} measures this form on CivicBench. \hfill$\square$

\paragraph{Linear growth and the break budget} The reach never decreases as the budget grows. It is a maximum over $\B_\eps(x)$, and $\B_\eps(x)$ only gains clips as $\eps$ grows. Suppose, in addition, that $r(x,\eps)=\eps\, r(x,1)$ for the budgets of interest, with $r(x,1)>0$. The condition $\rho(x)\geq r(x,\eps)$ of a generated clip then becomes $\eps\leq\rho(x)/r(x,1)$, and the condition $\rho(x)>r(x,\eps)$ of a real clip becomes $\eps<\rho(x)/r(x,1)$. The clip thus keeps its label up to the budget $\rho(x)/r(x,1)$, its break budget. Monotonicity is proven above. Linear growth is an empirical regularity, which \cref{supp:theory} measures.

\paragraph{\cref{prop:ties}} The proof bounds how far a perturbation can move each cost of block matching. Let $a$ be a block of the current frame with $n$ pixels $i$, and let $b_{\cdot,d}$ be its reference block displaced by $d$. The cost of candidate displacement $d$ is $c_d=\frac{1}{n}\sum_i|a_i-b_{i,d}|$. Perturbing $a$ and $b$ by at most $\eps$ per pixel changes each term $|a_i-b_{i,d}|$ by at most $2\eps$, so each cost changes by at most $2\eps$. The difference between any two costs then changes by at most $4\eps$. When $g>4\eps$, the lowest cost therefore stays below every other cost, and the motion vector is unchanged. For the flip statement, let $d_1$ be the chosen displacement and $d_2$ the second best, so that $g=c_{d_2}-c_{d_1}$. Let $P$ be the set of pixels whose value $a_i$ lies between $b_{i,d_1}$ and $b_{i,d_2}$ and at least $\eps$ from $b_{i,d_2}$, so that $q=|P|/n$. The attack moves each $a_i$ with $i\in P$ by $\eps$ toward $b_{i,d_2}$ and changes no other pixel. On each such pixel, $|a_i-b_{i,d_2}|$ falls by exactly $\eps$ and $|a_i-b_{i,d_1}|$ rises by exactly $\eps$. So $c_{d_2}$ falls and $c_{d_1}$ rises, each by exactly $\eps q$. The gap thus shrinks by $2\eps q$, and $d_2$ becomes cheaper than $d_1$ when $g<2\eps q$. Conversely, changing the current frame alone moves each cost by at most $\eps$ and the difference of two costs by at most $2\eps$. No block with $g>2\eps$ can therefore be flipped this way. At $g=2\eps$, the attack can at most create a tie, which the codec resolves by its tie-breaking rule. \hfill$\square$

\paragraph{\cref{prop:align}} Each part differentiates a loss with respect to a score. The size of this derivative is the push, and a gradient step moves the score against the derivative's sign. (i)~For a generated clip, cross-entropy on $x+\delta$ is $-\log\sigma(z')$. Its derivative with respect to $z'$ is $-\sigma(-z')$, so the push on $z'$ is $\sigma(-z')$, and a step raises $z'$. (ii)~Write $p=\sigma(z)$ and $p'=\sigma(z')$. The Kullback-Leibler (KL) divergence of the TRADES term is then $\mathrm{KL}(p\,\|\,p')=p\log(p/p')+(1-p)\log\big((1-p)/(1-p')\big)$. Since $dp'/dz'=p'(1-p')$, its derivative with respect to $z'$ is $-p(1-p')+(1-p)p'=p'-p$. This derivative vanishes at $z'=z$ and has the sign of $z'-z$. With the weight $\eta$, the term therefore pushes $z'$ toward $z$ by $\eta|p-p'|$. The logistic function has slope $\sigma'\leq\frac14$, so also $|p-p'|\leq|z-z'|/4$. The derivative of the KL divergence with respect to $p$ is $\log\frac{p}{1-p}-\log\frac{p'}{1-p'}=z-z'$. Its derivative with respect to $z$ is therefore $\sigma'(z)(z-z')$, and a step moves $z$ toward $z'$. The term is nonnegative and zero exactly when $z'=z$. The threshold appears nowhere in it, so this holds wherever $z$ sits. (iii)~Write $\rho$ for $\rho(x+\delta)$, measured at $\hat\tau$, and let $u=(\kappa-\rho)/\kappa$. The derivative of $\gamma\,\mathrm{softplus}(u)$ with respect to $\rho$ is $-\gamma\sigma(u)/\kappa$, so a step raises $\rho$. For $\rho\leq\kappa$, $u\geq0$ and $\sigma(u)\geq\frac12$, so the push is at least $\gamma/(2\kappa)$, whatever $\hat\tau$ is. Moreover, $\mathrm{softplus}(u)\geq\log2\cdot\mathbf{1}[u\geq0]$, where $\mathbf{1}[\cdot]$ is 1 if its condition holds and 0 otherwise. Summed over the attacked fakes, the term is thus at least $\gamma\log2$ times the number of attacked fakes with $\rho\leq\kappa$. Since $\rho$ is the margin after the training attack, these are the fakes that this attack breaks or leaves with a margin of at most $\kappa$ at $\hat\tau$. The softplus is convex. Frank and Niles-Weed~\cite{frank2023adversarial} show that convex surrogate losses are not consistent for adversarial risk, so minimizing them need not minimize the error under attack. Part (iii) therefore makes no claim of consistency. It concerns only the size of the push at the threshold. \hfill$\square$

\section{The Survival Condition, Measured}
\label{supp:theory}

\Cref{prop:survival} is proven (\cref{supp:proofs}), but it uses the true reach, and an attack finds only a lower bound on it. This section measures how well the proposition predicts survival under attack when the reach is estimated, and whether the reach grows linearly in the budget.

\paragraph{On CivicBench, both directions} In every CivicBench attack, we record the attacked score at each budget (\cref{supp:setup}). We can thus read the margin and the reach in score units without a further attack. The margin is the clean score's distance from CivicBench's own threshold at 5\% false alarms, on the side of the clip's label. The reach at a budget is how far projected gradient descent (PGD) moved the score toward the threshold. PGD finds one reachable clip, not necessarily the worst. The reach that PGD finds is therefore a lower bound on the true reach, and the predicted share of clips that keep their label is an upper bound on the true share.

\Cref{tab:mr_civic} shows that the reach grows almost linearly. For every detector on \dclip, the median ratio of a clip's reach at 4 gray levels to its reach at 1 is 3.8 to 4.4. This linear growth turns one measurement into a break budget. We predict that a clip breaks at 4 gray levels when four times its reach at 1 exceeds its margin. This prediction matches the measured share of clips broken at 4 gray levels to within 11 percentage points, for evasion and for framing. The rows of \stream in \cref{tab:mr_civic} come from extra attacks on the first 120 clips of each fixed attack set, on which \method is attacked (\cref{supp:setup}). \stream's clips break at a fraction of a gray level, so every attacked clip breaks at 1 gray level. Its break budgets below 1 gray level are therefore extrapolations from its reach at 1. They mean only that the smallest possible attack suffices. Its reach at 4 gray levels is therefore not measured.

Against its own version without the margin term, \method raises the break budget of its weakest clips under evasion, not the median. By \cref{prop:survival}, a margin term should help the weakest clips, those nearest the threshold, more than the typical clip. We therefore compare the 10th percentile of the break budget as well as its median. Unpaired quantiles, as in \cref{tab:mr_civic}, mislead here. Some of \method's attacked generated clips lie near the threshold, and the other detectors on \dclip do not flag them at all. These clips enter only \method's quantiles. \Cref{tab:mr_paired} therefore compares break budgets on the clips attacked against both detectors, with bootstrap intervals over clips. Against the same model without its margin term, \method raises the 10th percentile of the break budget under evasion from 1.9 to 3.0 gray levels ($p=.020$). It does not change the median significantly ($p=.965$). Against TRADES and MART the 10th percentile rises as well. Against Madry's objective it rises by +0.6, but not significantly ($p=.142$). Under framing, only the difference of the 10th percentiles against TRADES is significant (+1.0, $p=.027$). The probe of Mavali et al.\ on \dclip has the largest break budgets on real clips, as expected from its small reach.

\begin{table}[t]
\centering
\caption{\textbf{Margin against reach on CivicBench}, PGD. Number $n$ of attacked clips, 10th percentile and median break budget $\rho/r(x,1)$ in gray levels (margin and reach in score units), median growth of the reach $r(4)/r(1)$, and share broken at 4 gray levels, measured and predicted from $r(x,1)$. --, not applicable, since every \stream clip breaks below 1 gray level.}
\label{tab:mr_civic}
\footnotesize
\setlength{\tabcolsep}{1.7pt}
\begin{tabular}{l l c cc c cc}
\toprule
& & & \multicolumn{2}{c}{break budget} & & \multicolumn{2}{c}{broken at 4}\\
\cmidrule(lr){4-5}\cmidrule(lr){7-8}
& Detector & $n$ & 10\% & median & $r(4)/r(1)$ & measured & predicted\\
\midrule
\multirow{7}{*}{\rotatebox{90}{evasion}}
& \method & 196 & 1.5 & 6.8 & 4.0 & 24\% & 25\%\\
& w/o margin term & 108 & 1.9 & 7.6 & 4.2 & 28\% & 25\%\\
& Madry & 179 & 2.0 & 7.3 & 4.2 & 29\% & 23\%\\
& TRADES & 106 & 1.9 & 7.3 & 4.2 & 25\% & 23\%\\
& MART & 103 & 1.8 & 7.0 & 4.2 & 28\% & 28\%\\
& Mavali, \dclip & 27 & $-$0.30 & 5.9 & 4.2 & 41\% & 41\%\\
& \stream & 22 & 0.06 & 0.08 & -- & 100\% & 100\%\\
\midrule
\multirow{7}{*}{\rotatebox{90}{framing}}
& \method & 170 & 2.9 & 10.6 & 4.2 & 15\% & 14\%\\
& w/o margin term & 97 & 3.0 & 10.5 & 4.0 & 16\% & 14\%\\
& Madry & 171 & 2.5 & 9.2 & 4.0 & 17\% & 16\%\\
& TRADES & 100 & 2.0 & 9.4 & 4.0 & 22\% & 21\%\\
& MART & 96 & 3.6 & 10.6 & 4.0 & 12\% & 12\%\\
& Mavali, \dclip & 104 & 5.8 & 20.2 & 4.0 & 6\% & 6\%\\
& \stream & 106 & 0.09 & 0.26 & -- & 100\% & 100\%\\
\bottomrule
\end{tabular}
\end{table}

\begin{table}[t]
\centering
\caption{\textbf{Paired break budgets on CivicBench}, on the $n$ clips attacked against both \method (ours) and the other detector. The 10th percentile and median break budget in gray levels, and the difference of 10th percentiles, ours minus other, with its 95\% bootstrap interval.}
\label{tab:mr_paired}
\footnotesize
\setlength{\tabcolsep}{2.2pt}
\begin{tabular}{l l c cc c cc}
\toprule
& & & \multicolumn{3}{c}{10th percentile} & \multicolumn{2}{c}{median}\\
\cmidrule(lr){4-6}\cmidrule(lr){7-8}
& Other & $n$ & ours & other & difference & ours & other\\
\midrule
\multirow{5}{*}{\rotatebox{90}{evasion}}
& w/o margin & 108 & 3.0 & 1.9 & +1.2 {\scriptsize[+0.3, +2.1]} & 7.7 & 7.6\\
& Madry & 179 & 2.6 & 2.0 & +0.6 {\scriptsize[$-$0.3, +1.3]} & 7.5 & 7.3\\
& TRADES & 106 & 3.0 & 1.9 & +1.1 {\scriptsize[+0.3, +1.7]} & 7.7 & 7.3\\
& MART & 103 & 3.2 & 1.8 & +1.3 {\scriptsize[+0.7, +2.1]} & 7.8 & 7.0\\
& Mavali, \dclip & 27 & 5.3 & $-$0.30 & +5.6 {\scriptsize[+3.2, +11.1]} & 14.2 & 5.9\\
\midrule
\multirow{5}{*}{\rotatebox{90}{framing}}
& w/o margin & 95 & 3.9 & 3.6 & +0.3 {\scriptsize[$-$0.6, +1.9]} & 11.1 & 10.6\\
& Madry & 164 & 3.5 & 3.5 & +0.0 {\scriptsize[$-$0.8, +1.3]} & 10.8 & 9.3\\
& TRADES & 97 & 3.8 & 2.8 & +1.0 {\scriptsize[+0.1, +2.3]} & 11.0 & 9.6\\
& MART & 95 & 4.5 & 3.7 & +0.8 {\scriptsize[$-$0.3, +1.6]} & 11.1 & 10.7\\
& Mavali, \dclip & 97 & 3.3 & 5.8 & $-$2.5 {\scriptsize[$-$5.2, $-$0.3]} & 11.0 & 21.1\\
\bottomrule
\end{tabular}
\end{table}

\begin{figure}[t]
\centering
\includegraphics[width=\linewidth]{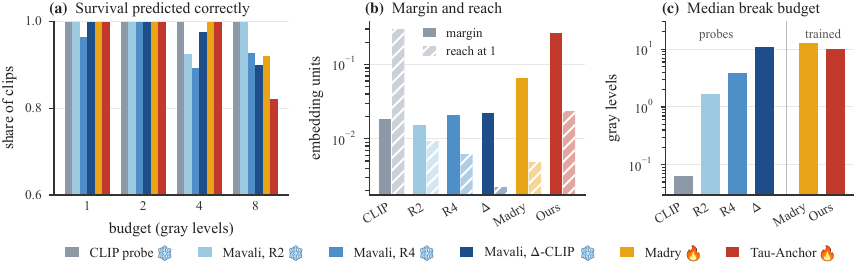}
\caption{\textbf{Margin against reach on \aigvd}, evasion, on the clips of \cref{tab:v1}. (a)~Share of clips for which \cref{prop:survival} correctly predicts survival, per budget and detector. (b)~Median margin and median reach at 1 gray level, in embedding units. (c)~Median break budget in gray levels.}
\label{fig:theory}
\end{figure}

\begin{figure}[t]
\centering
\includegraphics[width=0.55\linewidth]{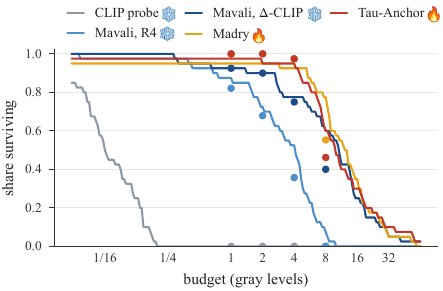}
\caption{\textbf{Survival predicted from one measurement.} Lines give the share of clips whose break budget $\rho/r(x,1)$ exceeds each budget. Dots give the measured share of clips that keep their label under attack. Both use the clips of \cref{tab:v1}.}
\label{fig:breakcdf}
\end{figure}

\paragraph{On \aigvd, evasion} Here too, the reach of \method grows about linearly in the budget. We estimate the reach in embedding units with 30 steps of PGD that maximize \cref{eq:disp}. Each detector flags 28 to 40 of the 40 attacked generated clips of fold 0, and we use the clips that it flags. \Cref{tab:v1} reports their margin, reach, and break budget. Per budget, it also reports the share of clips for which \cref{prop:survival} correctly predicts survival from the reach measured at that budget. In its rows, ``w/o averaging'' means without weight averaging. \Cref{fig:theory} plots these quantities, and \cref{fig:breakcdf} turns the break budget into the share of clips that keep their label. From 1 to 2 gray levels, \method's reach grows by a factor of 1.9. From 1 to 4 and 8 gray levels, it grows by 3.7 and 7.8. Up to 4 gray levels, nearly every clip of this sample keeps its label, so the prediction is easy there. At 8 gray levels, it is correct for 82.1\% of \method's clips. Its errors are all optimistic, predicting survival for clips that the attack breaks, because 30 steps of PGD underestimate the reach.

Linear probes on more robust encoders have larger break budgets (\cref{fig:theory}c). The probe of Mavali et al.\ on \dclip has about the same margin as the CLIP probe but a reach 132 times smaller. By \cref{prop:survival}, this smaller reach is the source of its robustness. The norm of the embedding shift is 6 to 7 times its component along the head $w$. Bounds based on the norm alone are therefore loose. The directional form of \cref{eq:disp} counts only the shift along $w$, which keeps the condition tight.

\begin{table}[t]
\centering
\caption{\textbf{Margin and reach per detector on \aigvd}, evasion. Medians of the margin $\rho$ and the reach $r(x,1)$, in embedding units, and of the break budget, in gray levels. Columns 1 to 8 give the share of clips for which \cref{prop:survival} correctly predicts survival at that budget in gray levels.}
\label{tab:v1}
\footnotesize
\setlength{\tabcolsep}{2.5pt}
\resizebox{0.55\linewidth}{!}{\begin{tabular}{l ccc cccc}
\toprule
Detector & $\rho$ & $r(x,1)$ & Break & 1 & 2 & 4 & 8\\
\midrule
CLIP probe & 0.019 & 0.2962 & 0.06 & 100\% & 100\% & 100\% & 100\%\\
Mavali, R2 & 0.016 & 0.0093 & 1.7 & 100\% & 100\% & 92.5\% & 100\%\\
Mavali, R4 & 0.021 & 0.0062 & 3.9 & 96.4\% & 100\% & 89.3\% & 92.9\%\\
Mavali, \dclip & 0.022 & 0.0022 & 10.7 & 100\% & 100\% & 97.5\% & 90\%\\
Madry & 0.066 & 0.0047 & 12.9 & 100\% & 100\% & 100\% & 92.1\%\\
\method w/o averaging & 0.334 & 0.0268 & 9.5 & 100\% & 100\% & 97.5\% & 92.5\%\\
\method & 0.268 & 0.0234 & 10.0 & 100\% & 100\% & 100\% & 82.1\%\\
\bottomrule
\end{tabular}}
\end{table}

\paragraph{Why only the last blocks} The comparison above shows that the encoder sets how far an attack can move the score. \method therefore changes as little of the encoder as needed for accuracy. It fine-tunes only the last 4 of the 32 transformer blocks of \dclip, so the first 28 keep their adversarial pretraining. Fine-tuning raises margin and reach together, compared with the probe of Mavali et al.\ on the frozen encoder. For \method, the median margin of flagged generated clips rises from 0.022 to 0.268 and the median reach at 1 gray level rises from 0.0022 to 0.0234. The median break budget thus stays near 10 gray levels (10.7 for Mavali et al.'s probe, 10.0 for \method, \cref{fig:theory}c). The gain of fine-tuning lies in clean accuracy (\cref{tab:ablation_aigvd}).

\section{Why Motion-Vector Detectors Fail}
\label{supp:mvties}

In real and generated clips alike, most blocks have a gap $g$ below $2\eps$, so the instability of motion vectors affects both classes. By \cref{prop:ties} (proof in \cref{supp:proofs}), a motion vector can flip only if the gap $g$ of its block is at most $4\eps$. Here we measure how common such small gaps are. We ran exhaustive block matching on the luma (brightness) channel. A codec approximates this search. Our search uses $16\times16$ blocks, integer displacements of up to 8 pixels, and the mean absolute difference as cost. We used 60 real and 60 generated clips from fold 0 of each of \aigvd and \gvb. From CivicBench, we used 200 real and 200 generated clips. We measured every block of every pair of consecutive frames. \Cref{tab:ties} reports the median gap between the lowest and second-lowest cost and the share of blocks guaranteed stable ($g>4\eps$). It also reports the share whose gap is below $2\eps$. Apart from exact ties at $2\eps$, only these blocks can flip when the attack changes the current frame alone (\cref{prop:ties}). The lower rows give the gap to the best candidate at least 2 pixels away from the chosen one, so a flip there moves the vector by 2 pixels or more. We measured it on \aigvd and \gvb only. The generated clips of CivicBench have somewhat more blocks with a gap below $2\eps$.

Codecs used in practice also compare blocks against references interpolated at subpixel positions. An interpolated sample is a fixed linear filter $h$ applied to reference pixels. Let $\norm{h}_1$ be the sum of the filter's absolute coefficients. A perturbation of at most $\eps$ per pixel then moves an interpolated sample by at most $\norm{h}_1\eps$. Each cost therefore moves by at most $(1+\norm{h}_1)\eps$, and the guarantee of \cref{prop:ties} becomes $g>2(1+\norm{h}_1)\eps$. H.264 interpolates with a 6-tap filter. Its $\norm{h}_1$ is $52/32$ at the horizontal and vertical half-sample positions and $(52/32)^2\approx2.64$ at the central one, where the filter is applied in both directions. The guarantee thus tightens from $g>4\eps$ to as much as $g>7.28\eps$, and fewer blocks are guaranteed stable. Codecs also smooth block edges in the reference frames, a step called in-loop filtering. These bounds leave it out. Our searches that need only scores run through the actual codec, with its interpolation and in-loop filtering, so their results already include both effects.

\paragraph{Gradient attacks and searches}\label{supp:mvattacks} We never report the robustness of a motion-vector detector from gradient attacks alone. Such a detector scores a clip only after re-encoding it. Block matching in the codec outputs a piecewise-constant motion field, whose gradient is zero almost everywhere (\cref{sec:mv}). A gradient attack must therefore differentiate a smooth stand-in for block matching, a model of the codec through which gradients can flow. Such a stand-in cannot follow the discrete vector flips of \cref{prop:ties}. A clip therefore counts as broken when any of our attacks breaks it, including the searches that need only scores. Under these attacks, \mvmob on \gvb is the one motion-vector detector that keeps most generated clips flagged, and its robustness does not carry over to \aigvd (\cref{tab:main}).

\section{Setup in Detail}
\label{supp:setup}

\begin{figure}[t]
\centering
\includegraphics[width=\linewidth]{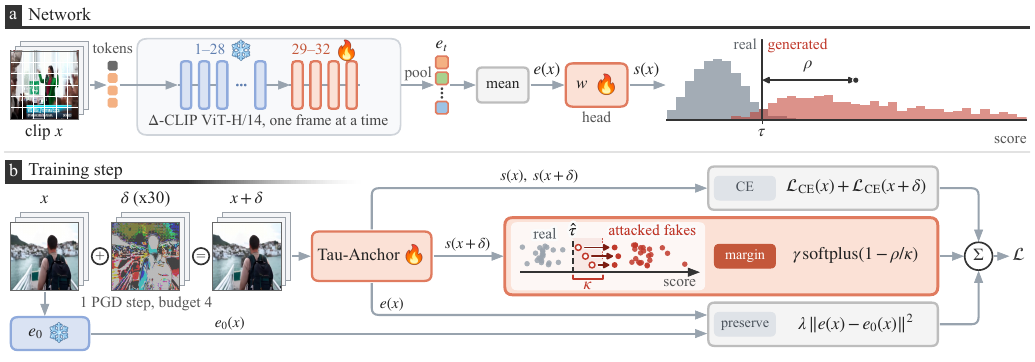}
\caption{\textbf{Architecture of \method.} (a) The frame embeddings $e_t$ of a CivicBench clip, the clip embedding $e(x)$ as their normalized mean, the head $w$, and the scores of the test clips of fold 0 of \aigvd with the threshold $\tau$ and the margin $\rho$ of one clip. (b) One training step, with $\delta$ shown 30 times larger.}
\label{fig:arch}
\end{figure}

\paragraph{Architecture} \Cref{fig:arch}a shows \method on a CivicBench clip of an invented politician. The \dclip encoder embeds each frame separately, with the same weights for every frame. The clip embedding $e(x)$ is the normalized mean of the normalized frame embeddings $e_t$. We train only the last four transformer blocks (29--32), the final normalization, the projection, and the head $w$. The decision boundary at the threshold $\tau$ is perpendicular to $w$. A clip's margin $\rho$ is its signed distance from this boundary.

\paragraph{Data} Detectors are trained and tested on subsets of two general benchmarks, \aigvd and \gvb. The generated clips of \aigvd come from six of its generators, which render a shared set of prompts. Its real clips come from its source videos. \gvb pairs four generators with real videos. \Cref{tab:data} lists the clips per source.

\paragraph{Preprocessing} Every clip, in CivicBench and in the general benchmarks, is prepared in the same three steps. First, it is re-encoded with x264, the open-source H.264 encoder, at its default quality setting. This setting is a constant rate factor (CRF) of 23. FaceForensics++~\cite{rossler2019faceforensics} uses the same value for its light compression. Key frames are at most 250 frames apart, and the source frame rate is kept. Second, the clip is scaled to $512\times512$ pixels, as in training. Scaling does not keep the aspect ratio. It squashes wide frames and stretches the 28 real portrait clips of CivicBench, which change shape most. Third, we sample 11 frames at about 8 frames per second from the start of the clip. For a video at $f$ frames per second, we take every $k$-th frame, with $k$ the integer nearest $f/8$.

\begin{table}[t]
\centering
\caption{\textbf{Stability of motion vectors.} Median gap $g$ in gray levels, share of blocks guaranteed stable ($g>4\eps$), and share with $g<2\eps$, the only blocks that an attack on the current frame alone can flip. Budgets $\eps$ are 1 and 2 gray levels. R marks real clips, G generated ones. The rows marked nearest give the gap to the second-best candidate, and the lower rows give the gap to the best candidate at least 2 pixels away.}
\label{tab:ties}
\small
\setlength{\tabcolsep}{2pt}
\begin{tabular}{l l c cc cc}
\toprule
& & & \multicolumn{2}{c}{Stable} & \multicolumn{2}{c}{Gap $<2\eps$}\\
\cmidrule(lr){4-5}\cmidrule(lr){6-7}
Candidates & Set & Gap & 1 & 2 & 1 & 2\\
\midrule
\multirow{6}{*}{nearest} & CivicBench R & 0.34 & 10.9\% & 4.7\% & 80\% & 89\%\\
& CivicBench G & 0.15 & 4.8\% & 1.7\% & 90\% & 95\%\\
& \aigvd R & 0.34 & 9.1\% & 3.0\% & 81\% & 91\%\\
& \aigvd G & 0.25 & 9.7\% & 3.3\% & 82\% & 90\%\\
& \gvb R & 0.15 & 6.3\% & 2.3\% & 88\% & 94\%\\
& \gvb G & 0.23 & 6.7\% & 2.3\% & 86\% & 93\%\\
\midrule
\multirow{4}{*}{$\geq$2 px away} & \aigvd R & 1.20 & 25.0\% & 11.2\% & 60\% & 75\%\\
& \aigvd G & 0.75 & 21.4\% & 10.5\% & 67\% & 79\%\\
& \gvb R & 0.49 & 15.5\% & 7.2\% & 74\% & 84\%\\
& \gvb G & 0.71 & 17.4\% & 7.6\% & 70\% & 83\%\\
\bottomrule
\end{tabular}
\end{table}

\paragraph{Folds} Each general benchmark is split into five folds grouped by prompt, the text that a generator was given. No prompt thus appears in both training and test data. The model for fold $k$ trains on the other four folds and is tested on fold $k$. Its threshold is set on the real clips of fold $k$.

\begin{table}[t]
\centering
\caption{\textbf{Data of the general benchmarks.} Clips per source. Each fold holds one fifth of every source, grouped by prompt.}
\label{tab:data}
\small
\setlength{\tabcolsep}{4pt}
\begin{tabular}{l r l r}
\toprule
\multicolumn{2}{c}{\aigvd} & \multicolumn{2}{c}{\gvb}\\
\cmidrule(lr){1-2}\cmidrule(lr){3-4}
Source & Clips & Source & Clips\\
\midrule
real videos & 10,000 & real videos (Vript~\cite{yang2024vript}) & 10,000\\
AccVideo & 830 & ModelScope & 1,250\\
CogVideoX1.5 & 830 & Pika & 1,250\\
HunyuanVideo & 830 & Text2Video-Zero & 1,250\\
Open-Sora & 830 & VideoCrafter2 & 1,250\\
Pyramid-Flow & 830 & & \\
Wan2.1 & 830 & & \\
\bottomrule
\end{tabular}
\end{table}

\paragraph{Generator weights} The loss weights $\omega_i$ of \cref{alg:train} include a weight for each generator. A generator's weight is the median margin of all generated training clips divided by the median margin of its own clips, clipped to the range 1 to 3. Generators whose clips sit closest to the threshold thus count up to three times as much in the loss. The margins come from out-of-fold scores of the linear probe that \method starts from, so no clip is scored by a probe trained on it. Only clips of the training folds are scored. Each of their probes, however, is fit on the four other folds, which include the test fold, so test labels enter the generator weights indirectly.

\paragraph{Training step} \Cref{fig:arch}b shows one training step of \method, and \cref{alg:train} lists it. The training clips come from the training folds of \aigvd or \gvb. CivicBench is used only for testing. We fixed every setting of \method on \aigvd, partly with its test folds, before CivicBench was built. One later variant, a margin on real clips, was tested on CivicBench under a rule fixed in advance and was not adopted (\cref{supp:ablation_aigvd}). Each step trains on clean clips and on attacked clips, whose perturbation $\delta$ stays within 4 gray levels. The loss combines cross-entropy (CE), the margin term of weight $\gamma$, and a preservation term of weight $\lambda$. The preservation term keeps the embedding of each sampled frame near that of the frozen encoder $e_0$. The margin term pushes up the attacked generated clips, most strongly those whose margin at the threshold estimate $\hat\tau$ is below $\kappa$. \Cref{tab:hparams} lists all training settings.

\begin{table}[t]
\centering
\caption{\textbf{Training settings of \method.} Step sizes and budgets are in gray levels. Madry's objective, TRADES, and MART change the loss, and TRADES and MART also drop the generator weights. KL, the Kullback-Leibler term of TRADES and MART.}
\label{tab:hparams}
\small
\setlength{\tabcolsep}{3pt}
\begin{tabular}{l l}
\toprule
Setting & Value\\
\midrule
Encoder & \dclip ViT-H/14, 336 px\\
Trained parts & last 4 blocks, norm, projection, head\\
Frames per clip & 2 random of 11\\
Batch size & 4 clips\\
Epochs & 1.5\\
Learning rate & $10^{-5}$ encoder, $10^{-3}$ head\\
Optimizer & AdamW, weight decay $10^{-4}$\\
Training attack & random start, 1 step of 2, budget 4\\
Margin & $\kappa=0.15$, $\gamma=1$\\
Threshold estimate & every 100 steps\\
Preservation weight & $\lambda=1$\\
Generator weights & 1 to 3, from training-fold margins\\
KL weight & $\eta=6$, TRADES and MART\\
Weight averaging & exponential, decay 0.999\\
Precision & 16-bit mixed\\
\bottomrule
\end{tabular}
\end{table}

\paragraph{Baselines} Mavali et al.~\cite{mavali2026adversarial} train a linear probe on the frame embeddings of a robust CLIP encoder, R2 or R4. These are the ViT-L/14 encoders of Schlarmann et al.~\cite{schlarmann2024robust}, fine-tuned adversarially at 2 and 4 gray levels. We train the probe on clip embeddings, the mean of the frame embeddings. We also apply the recipe to \dclip. Shahjahan et al.~\cite{shahjahan2026improving} feed CLIP features and a denoising residual to a small attention head. The residual is the difference between a frame and its denoised copy. Their released code has no trained weights, so we retrain the head on our folds with their denoiser and feature extractor. The other published detectors are trained on our folds with their released code. D3~\cite{zheng2025d3} needs no training and is used as released. It scores a clip by the spread of the second-order differences of its frame features. We attack one or two detectors per kind of signal rather than every published detector. DeMamba~\cite{chen2024demamba} and UNITE~\cite{kundu2025towards} read frame features over time, as \restrav and D3 do, and are not built to resist attacks. The anytime detector of Cakiroglu et al.~\cite{cakiroglu2026detect} reads the compressed bitstream, as \stream and \codecmv do (\cref{supp:attacks}).

\paragraph{Adversarial-training baselines} Madry's objective, TRADES, and MART are standard losses of adversarial training (\cref{supp:civic}). They replace \method's loss and keep its architecture, initialization, training attack, budget, number of steps, and weight averaging. Madry's objective keeps \method's generator weights, while TRADES and MART drop them.

\paragraph{Attacks} All attacks run on the full pipeline, from the 8-bit clip to the score. The attack changes the 11 processed frames. Frame-based detectors score these 8-bit frames as saved. Detectors that read the encoded stream or its motion vectors first encode them with the codec they read. \Cref{supp:attacks} also re-encodes every successful attack. \method and the CLIP-based detectors get exact gradients. Inference is deterministic, so there is no need to average gradients over random transformations (expectation over transformation). PGD (projected gradient descent) repeats two operations. It moves every pixel a step against the gradient of the score for evasion, or along it for framing, and then clips the change back into the budget. Each step follows the sign of the score's gradient, so its size does not depend on where the threshold lies. It runs 100 steps, each a quarter of the budget. Every 10 steps, it checks whether the saved 8-bit clip crosses the threshold. APGD (Auto-PGD) adapts its step size during the run. It follows Croce and Hein~\cite{croce2020reliable}, with 100 steps and 5 restarts. Square~\cite{andriushchenko2020square} changes random square patches and keeps a change only if the score drops for evasion, or rises for framing. It thus needs only scores. It uses 1,000 queries, each a scored candidate, with the standard schedule of patch sizes.

\begin{table}[t]
\centering
\caption{\textbf{Confidence intervals for CivicBench in \cref{tab:main}.} Each cell gives the value and its 95\% bootstrap interval. The evasion columns give the TPR under evasion, and the framing columns give the attacked false-alarm rate, at 1, 2, and 4 gray levels. Evasion cells rest on 16 to 225 attacked clips per detector, and framing cells on 69 to 185 (\cref{supp:setup}).}
\label{tab:ci_civic}
\scriptsize
\setlength{\tabcolsep}{3pt}
\begin{tabular}{l ccc ccc}
\toprule
& \multicolumn{3}{c}{evasion $\uparrow$} & \multicolumn{3}{c}{framing $\downarrow$}\\
\cmidrule(lr){2-4}\cmidrule(lr){5-7}
Detector & 1 & 2 & 4 & 1 & 2 & 4\\
\midrule
CLIP probe & .000 {\tiny[.000, .000]} & .000 {\tiny[.000, .000]} & .000 {\tiny[.000, .000]} & 1.000 {\tiny[1.000, 1.000]} & 1.000 {\tiny[1.000, 1.000]} & 1.000 {\tiny[1.000, 1.000]}\\
\restrav & .000 {\tiny[.000, .000]} & .000 {\tiny[.000, .000]} & .000 {\tiny[.000, .000]} & .986 {\tiny[.955, 1.000]} & 1.000 {\tiny[1.000, 1.000]} & 1.000 {\tiny[1.000, 1.000]}\\
D3 & .000 {\tiny[.000, .000]} & .000 {\tiny[.000, .000]} & .000 {\tiny[.000, .000]} & 1.000 {\tiny[1.000, 1.000]} & 1.000 {\tiny[1.000, 1.000]} & 1.000 {\tiny[1.000, 1.000]}\\
\stream & .000 {\tiny[.000, .000]} & .000 {\tiny[.000, .000]} & .000 {\tiny[.000, .000]} & 1.000 {\tiny[1.000, 1.000]} & 1.000 {\tiny[1.000, 1.000]} & 1.000 {\tiny[1.000, 1.000]}\\
\streammv & .018 {\tiny[.010, .027]} & .013 {\tiny[.006, .021]} & .011 {\tiny[.004, .019]} & .768 {\tiny[.667, .870]} & .783 {\tiny[.681, .882]} & .812 {\tiny[.710, .899]}\\
\mvmob & .045 {\tiny[.035, .054]} & .043 {\tiny[.033, .053]} & .035 {\tiny[.026, .044]} & .348 {\tiny[.232, .464]} & .493 {\tiny[.371, .609]} & .565 {\tiny[.448, .681]}\\
Mavali, R2 & .236 {\tiny[.195, .277]} & .164 {\tiny[.125, .205]} & .135 {\tiny[.098, .176]} & .261 {\tiny[.159, .373]} & .536 {\tiny[.418, .662]} & .884 {\tiny[.806, .956]}\\
Mavali, R4 & .339 {\tiny[.309, .365]} & .289 {\tiny[.249, .328]} & .239 {\tiny[.199, .281]} & .217 {\tiny[.127, .314]} & .304 {\tiny[.200, .418]} & .623 {\tiny[.500, .739]}\\
Mavali, \dclip & .135 {\tiny[.109, .160]} & .130 {\tiny[.102, .156]} & .099 {\tiny[.068, .128]} & .046 {\tiny[.009, .083]} & .046 {\tiny[.009, .083]} & .101 {\tiny[.046, .164]}\\
Shahjahan et al. & .023 {\tiny[.000, .061]} & .000 {\tiny[.000, .000]} & .000 {\tiny[.000, .000]} & .913 {\tiny[.838, .971]} & .971 {\tiny[.926, 1.000]} & .986 {\tiny[.955, 1.000]}\\
\method & .731 {\tiny[.705, .755]} & .668 {\tiny[.631, .703]} & .585 {\tiny[.540, .627]} & .124 {\tiny[.081, .170]} & .141 {\tiny[.092, .191]} & .222 {\tiny[.163, .283]}\\
\bottomrule
\end{tabular}
\end{table}

\paragraph{Attacks on motion-vector detectors} These detectors get gradients only through smooth stand-ins (\cref{supp:mvties}). The stand-ins are the optical-flow network RAFT~\cite{teed2020raft} for \mvmob and \streammv and a soft motion search for \codecmv, a detector of the audit toolkit that reads codec motion vectors (\cref{supp:attacks}). These detectors also face searches that need only scores, each with 1,000 queries per budget. The searches are Square and, for \mvmob and \codecmv, SimBA~\cite{guo2019simple}, the simple black-box attack. SimBA changes one coefficient of a low-frequency cosine basis at a time.

\paragraph{Budgets and coverage} Attacks run separately at each budget. A clip counts as broken at the lowest budget at which any attack succeeds. On CivicBench, the published detectors and those built to resist attacks face PGD, APGD, and Square on all their attacked clips. \method, Madry's objective, and the ablations face PGD on all their attacked clips. \method and Madry's objective also face APGD at 4 gray levels on all their paired clips except the 29 Mochi clips, which joined the sets later. Wherever APGD broke a clip at 4 gray levels that PGD had not broken at a lower budget, APGD also ran at 1 and 2 gray levels. Both also face APGD and Square at 1, 2, and 4 gray levels on 20 clips per direction (evasion and framing).

\paragraph{Thresholds after the real-clip fix}\label{supp:thresholds} Replacing the 131 title cards and duplicate clips (\cref{supp:civic}) moved the 5\% threshold of each detector. It fell most for published detectors that scored title cards as generated. The attacks ran at the earlier thresholds and stopped at their first success. We therefore judge each attacked clip from its recorded scores at the current threshold. An attack that stopped between the two thresholds stays undecided. It counts as failed for every detector except \method, so each comparison stays conservative for \method. Undecided evasions occur only for \streammv, \mvmob, and Mavali et al.\ on R2 and R4, whose TPRs under evasion are thus upper bounds.

\paragraph{Shahjahan et al.\ on CivicBench} This detector scores the same clip differently on A100 and V100 GPUs. On 100 random CivicBench clips, the two GPUs give different verdicts for 5\% of them, although their scores correlate at .989. Every saved PGD attack keeps its outcome on both GPUs, all 45 framings and the 4 evasions of clips that both GPUs flag. Its noise branch takes products of four copies of a noise residual, which magnify rounding errors. Our scoring and attacks ran on both kinds of GPU, so its attacked scores and its 5\% threshold do not share one scale. Its scores, however, fall into two clusters, and PGD moves an attacked clip from one cluster to the other. Its attacked rates therefore barely depend on the threshold. At thresholds that flag 2\% to 10\% of real clips, framing at 1 gray level reaches 90\% to 94\% of the attacked real clips, and 96\% to 99\% at 2 gray levels. Evasion succeeds on 90\% to 100\% of its flagged generated clips at every budget. \Cref{tab:main} reports the rates at the 5\% threshold.

\paragraph{Attack sets on CivicBench} Evasion attacks target flagged generated clips and framing attacks target real clips, at 1, 2, and 4 gray levels. \method and Madry's objective are attacked on the same fixed sets. The evasion set holds 225 generated clips that \method flags, about 28 per generator, listed so that generators alternate. The framing set holds 185 real clips, drawn from the four kinds of real scenes. Before any attack, \method flags 8\% of them, above 5\%, partly because they overrepresent legislatures, where clean false alarms are most common (\cref{fig:depth}b). Mavali et al.\ on \dclip and the other variants of \method are attacked on the first 120 clips of each set. The remaining detectors are attacked on up to 10 of their own flagged clips per generator and on 69 real clips. Framing attacks aim at the stricter threshold for 1\% false alarms, but a real clip counts as framed at a budget once any attack has reached the 5\% threshold at that budget or below. Every attack records the attacked score at each budget. \Cref{supp:theory} uses these scores to measure the reach.

\begin{algorithm}[t]
\caption{One training step of \method}
\label{alg:train}
\small
\begin{algorithmic}[1]
\Require batch $\{(x_i,y_i)\}_{i=1}^{N}$ with $y_i=1$ for the $N_g$ generated clips, and loss weights $\omega_i$ that
  balance the classes and weight the generators; encoder $e$ and head $(w,b)$ with trainable weights $\theta$; frozen
  copy $e_0$; averaged weights $\bar\theta$; threshold estimate $\hat\tau$; step count $t$; $\ell$ is cross-entropy
\State Sample 2 of the 11 frames of each clip
\State $\delta_i\sim\mathcal{U}[-\eps,\eps]$ per pixel \Comment{random start, $\eps=4$ gray levels}
\State $\delta_i\gets\mathrm{clip}_{[-\eps,\eps]}\big(\delta_i+\tfrac{\eps}{2}\,\mathrm{sign}\,\nabla_{\delta_i}\ell(x_i+\delta_i,y_i)\big)$
\State Clip every $x_i+\delta_i$ to the pixel range $[0,255]$
\State $\mathcal{L}\gets\frac{1}{N}\sum_i\omega_i\big[\ell(x_i,y_i)+\ell(x_i+\delta_i,y_i)\big]$
\State $\mathcal{L}\gets\mathcal{L}+\frac{1}{N_g}\sum_{i:\,y_i=1}\omega_i\,\mathcal{L}_\rho(x_i+\delta_i)$ \Comment{\cref{eq:margin}, $\hat\tau$ held fixed}
\State $\mathcal{L}\gets\mathcal{L}+\frac{\lambda}{N}\sum_i\norm{e(x_i)-e_0(x_i)}^2$ \Comment{preservation}
\State $\theta\gets\mathrm{AdamW}\big(\theta,\nabla_\theta\mathcal{L}\big)$
\State $\bar\theta\gets0.999\,\bar\theta+0.001\,\theta$ \Comment{weight averaging}
\If{$t\bmod100=0$}
  \State $\hat\tau\gets$ 95th percentile of $s$ on 256 real training clips
\EndIf
\Ensure the averaged weights $\bar\theta$, which are deployed
\end{algorithmic}
\end{algorithm}

\paragraph{Choice of attack sets} Two choices shape these sets. First, only generated clips that \method flags are attacked, which is conservative for \method. Of the 10,800 generated clips, \method flags 8,427. Of these, Madry's objective misses 581. Madry's objective flags 264 that \method misses. These clips sit just above the threshold of Madry's objective. Their median margin is 0.16 standard deviations of its real scores, against 1.54 on the clips that both flag. Leaving them out thus spares Madry's objective the clips that are easiest to break. Second, the sets take about 28 flagged generated clips from each generator, although generators differ in TPR. As a check, we compute each generator's TPR under evasion. It is the generator's clean TPR times its share of attacked clips that stay flagged. We then weight each generator's rate by its share of all generated clips. This moves \method's TPR under evasion at 4 gray levels from .585 to .604. It moves that of Madry's objective from .537 to .546. The ranking is unchanged. Over all detectors, this weighting changes a TPR under evasion by at most .029. Clean TPR covers all eight generators. Mochi's clips were added to the attack sets after the other generators had been attacked. The ablations of \cref{tab:ablation} use the first 120 clips of each set, which hold no Mochi clips.

\paragraph{Attack sets on the general benchmarks} On \aigvd and \gvb, we attack 40 flagged generated clips per benchmark and fold, alternating between generators, at up to 8 gray levels. In paired comparisons, every model is attacked on \method's clip set, each at its own threshold.

\paragraph{Clips that a detector cannot score} \mvmob extracts no motion vectors from 62 clips of \aigvd and 848 of \gvb. Most of them are real clips or static clips of Text2Video-Zero. We evaluate \mvmob on the remaining clips. Every other detector of \cref{tab:main} scores every clip. At the threshold of its attack run, D3 flagged none of Mochi's clips, so its attack set holds no Mochi clips.

\begin{table}[t]
\centering
\caption{\textbf{Confidence intervals for the general benchmarks in \cref{tab:main}.} 95\% bootstrap intervals of the clean AUC and TPR and of the TPR under evasion at 2 and 4 gray levels. At 2 and 4 gray levels, \stream keeps no attacked clip, so its intervals are [0, 0]. $^\ddagger$Trained on fold 0 only.}
\label{tab:ci}
\scriptsize
\setlength{\tabcolsep}{3pt}
\begin{tabular}{l cccc c cccc}
\toprule
& \multicolumn{4}{c}{\aigvd} && \multicolumn{4}{c}{\gvb}\\
\cmidrule{2-5}\cmidrule{7-10}
Method & AUC & TPR & 2 gray levels & 4 gray levels && AUC & TPR & 2 gray levels & 4 gray levels\\
\midrule
\stream & [.984, .987] & [.922, .939] & [0, 0] & [0, 0] && [.994, .996] & [.975, .983] & [0, 0] & [0, 0]\\
CLIP probe & [.953, .959] & [.757, .794] & [.000, .000] & [.000, .000] && [.996, .997] & [.985, .992] & [.000, .000] & [.000, .000]\\
Mavali, R2 & [.949, .956] & [.759, .792] & [.260, .506] & [.058, .240] && [.991, .993] & [.959, .972] & [.708, .918] & [.169, .456]\\
Mavali, R4 & [.948, .955] & [.767, .802] & [.530, .636] & [.279, .396] && [.986, .989] & [.939, .955] & [.807, .892] & [.601, .723]\\
Mavali, \dclip & [.937, .945] & [.698, .736] & [.586, .666] & [.526, .620] && [.987, .990] & [.936, .953] & [.916, .949] & [.828, .900]\\
Shahjahan et al.$^\ddagger$ & [.995, .998] & [1.000, 1.000] & [.000, .000] & [.000, .000] && [.684, .724] & [.190, .259] & [.000, .000] & [.000, .000]\\
Madry & [.985, .988] & [.923, .942] & [.900, .935] & [.832, .898] && [.999, .999] & [.996, .999] & [.996, .999] & [.996, .999]\\
TRADES & [.986, .989] & [.924, .943] & [.880, .928] & [.821, .892] && [.999, .999] & [.996, .999] & [.996, .999] & [.982, .999]\\
MART & [.987, .989] & [.930, .947] & [.912, .942] & [.849, .911] && [.999, .999] & [.996, .999] & [.996, .999] & [.982, .999]\\
\midrule
\method & [.986, .989] & [.925, .944] & [.917, .942] & [.852, .912] && [.999, .999] & [.996, .999] & [.982, .999] & [.982, .999]\\
\bottomrule
\end{tabular}
\end{table}

\section{Confidence Intervals, Paired Tests, and Five Folds}
\label{supp:ci}
\label{supp:folds}

\paragraph{Statistics} We report 95\% confidence intervals (CIs) from a bootstrap with 2,000 resamples. Real and generated clips are resampled separately. Detectors attacked on the same clips are compared paired, on the same resampled clips.

\paragraph{Intervals on CivicBench} Clean intervals resample prompts for generated clips and source videos for real clips, so clips that share a prompt or a source video are drawn together. \Cref{tab:civic_src} gives these clustered intervals for the clean AUC. \Cref{tab:ci_civic} gives the intervals of the attacked rates at 1, 2, and 4 gray levels. For the TPR under evasion, the bootstrap resamples the attacked generated clips by prompt and holds the clean TPR fixed. For the attacked false-alarm rate, it resamples the attacked real clips by source video.

\paragraph{Intervals on the general benchmarks} The clean resample draws the test clips of each fold with replacement, and every method is scored on the same draw. The attacked resample draws the attacked clips of each fold. Each resampled TPR under evasion is the clean TPR of a draw times the share of resampled attacked clips that stay flagged. Five baselines are attacked on \method's clips over five folds. They are Madry's objective, TRADES and MART, each with weight averaging, and Mavali et al.\ on R4 and on \dclip. The other detectors are attacked on their own flagged clips of fold 0, as in \cref{tab:main}. \Cref{tab:ci} gives the intervals of the clean AUC and TPR over five folds. It also gives those of the TPR under evasion at 2 and 4 gray levels, the attacked columns of \cref{tab:main}.

\begin{figure}[t]
\centering
\includegraphics[width=0.55\linewidth]{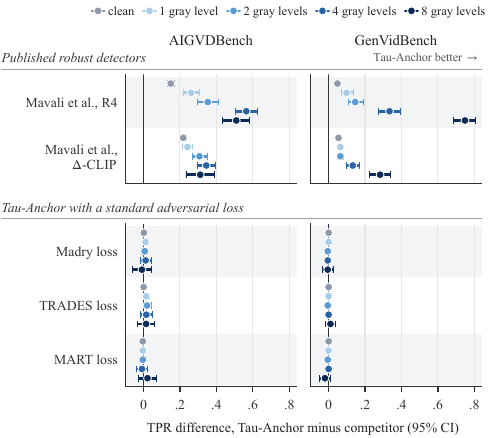}
\caption{\textbf{Paired differences.} TPR of \method minus each competitor's over five folds, clean and at 1 to 8 gray levels, with 95\% bootstrap intervals. Top, Mavali et al.\ on R4 and \dclip. Bottom, Madry's objective, TRADES, and MART. Positive differences favor \method.}
\label{fig:sig}
\end{figure}

\paragraph{Paired differences} Replacing \method's loss with a standard adversarial loss changes the TPR little. \Cref{tab:paired_ci} gives the paired differences behind \cref{fig:sig}. For the detectors attacked on \method's clips, the differences are paired. An interval entirely above zero means that \method is significantly better. The detector of Shahjahan et al.\ is attacked on its own clips and compared unpaired. In \cref{fig:sig}, the top panel shows the detectors of Mavali et al.\ and the bottom panel shows the standard adversarial losses. These losses replace \method's loss but keep its encoder, fine-tuning, and weight averaging.

\paragraph{Results on the general benchmarks} On the general benchmarks, too, \method keeps the most attacked fakes flagged (\cref{tab:main}). Under attack, the published detectors lose almost every generated clip at 1 or 2 gray levels. The exception is \mvmob on \gvb, with a TPR under evasion of .720 at 2 gray levels. The recipe of Mavali et al.\ on \dclip uses \method's own encoder. Against it, \method's TPR under evasion at 4 gray levels is higher by +.344 on \aigvd and +.133 on \gvb, paired over five folds. Every interval excludes zero. Without attack, \method's TPR over five folds exceeds \stream's by +.005 [$-$.007, +.016] on \aigvd. On \gvb it exceeds it by +.019 [+.014, +.023].

\begin{table}[t]
\centering
\caption{\textbf{Paired differences on the general benchmarks.} TPR of \method minus each competitor's, clean and at 1 to 8 gray levels, with 95\% bootstrap intervals. The Folds column gives the folds of the attacked comparison.}
\label{tab:paired_ci}
\scriptsize
\setlength{\tabcolsep}{2.5pt}
\resizebox{\linewidth}{!}{\begin{tabular}{l l l ccccc}
\toprule
Competitor & Set & Folds & Clean & 1 gray level & 2 gray levels & 4 gray levels & 8 gray levels\\
\midrule
Madry & \aigvd & 5 folds & +.002 [$-$.004, +.009] & +.012 [$-$.001, +.027] & +.007 [$-$.010, +.024] & +.012 [$-$.018, +.042] & $-$.008 [$-$.060, +.045]\\
Madry & \gvb & 5 folds & +.001 [$-$.001, +.002] & +.001 [$-$.001, +.002] & $-$.004 [$-$.016, +.001] & $-$.004 [$-$.016, +.001] & $-$.004 [$-$.034, +.024]\\
\addlinespace[2pt]
TRADES & \aigvd & 5 folds & +.001 [$-$.006, +.008] & +.015 [$-$.001, +.034] & +.020 [+.002, +.041] & +.015 [$-$.015, +.047] & +.014 [$-$.031, +.059]\\
TRADES & \gvb & 5 folds & +.001 [$-$.001, +.002] & +.001 [$-$.001, +.002] & $-$.004 [$-$.015, +.001] & +.001 [$-$.001, +.002] & +.011 [$-$.016, +.040]\\
\addlinespace[2pt]
MART & \aigvd & 5 folds & $-$.004 [$-$.010, +.003] & $-$.004 [$-$.010, +.003] & $-$.004 [$-$.017, +.011] & $-$.008 [$-$.040, +.023] & +.022 [$-$.025, +.071]\\
MART & \gvb & 5 folds & +.000 [$-$.001, +.001] & +.000 [$-$.001, +.001] & $-$.005 [$-$.016, +.001] & +.000 [$-$.001, +.001] & $-$.020 [$-$.051, +.011]\\
\addlinespace[2pt]
Mavali, \dclip & \aigvd & 5 folds & +.218 [+.200, +.235] & +.241 [+.215, +.269] & +.307 [+.268, +.348] & +.344 [+.295, +.393] & +.311 [+.233, +.392]\\
Mavali, \dclip & \gvb & 5 folds & +.055 [+.045, +.062] & +.065 [+.049, +.083] & +.065 [+.049, +.083] & +.133 [+.097, +.171] & +.283 [+.225, +.341]\\
\addlinespace[2pt]
Mavali, R4 & \aigvd & 5 folds & +.150 [+.134, +.167] & +.261 [+.222, +.309] & +.353 [+.298, +.409] & +.564 [+.504, +.628] & +.509 [+.435, +.580]\\
Mavali, R4 & \gvb & 5 folds & +.049 [+.042, +.058] & +.099 [+.072, +.134] & +.146 [+.108, +.192] & +.335 [+.272, +.396] & +.748 [+.686, +.808]\\
\addlinespace[2pt]
Shahjahan et al. & \aigvd & fold 0 & $-$.043 [$-$.061, $-$.028] & +.957 [+.939, +.972] & +.957 [+.939, +.972] & +.932 [+.874, +.968] & +.540 [+.390, +.681]\\
Shahjahan et al. & \gvb & fold 0 & +.774 [+.736, +.805] & +.996 [+.990, +.999] & +.996 [+.990, +.999] & +.996 [+.990, +.999] & +.896 [+.796, +.973]\\
\bottomrule
\end{tabular}}
\end{table}

\paragraph{Paired tests} Two tests compare models attacked on the same clips. McNemar's test uses the clips lost by exactly one of the two models. It tests whether one model loses more of them than chance allows. The Wilcoxon signed-rank test compares, clip by clip, how deep the attack pushed each model's score. Depth is measured by the catch rate, the lowest false-alarm rate at which the attacked clip would still be flagged. The deeper the clip falls, the higher its catch rate. The catch rate does not depend on the scale of the model's scores. On two training runs of the same recipe, the Wilcoxon test found no difference at any budget. However, the two runs lost different clips. At 8 gray levels, one run lost six clips that the other kept, and the other lost one clip that the first kept. The Wilcoxon test is thus stable across reruns, and we report it beside the bootstrap intervals and McNemar's test.

With weight averaging, \method and Madry's objective perform about equally on \aigvd. At 4 gray levels over all five folds of \aigvd, \method keeps six attacked clips that Madry's objective loses. Madry's objective keeps four that \method loses (McNemar $p=.75$). The clean TPRs differ by +.002 (95\% CI [$-$.004, +.009]).

\paragraph{Five folds} \Cref{tab:folds} gives means over the five folds, and \cref{tab:paired} gives the paired tests. Each benchmark has 200 attacked clips, 40 per fold. All three models of \cref{tab:folds} are attacked on these same clips. Madry's objective here uses weight averaging. \method and Madry's objective both flag 195 of them on \aigvd and 200 on \gvb. The TPR under evasion is the clean TPR times the share of attacked clips still flagged, computed per fold.

With weight averaging, the two objectives lose about as many clips at every budget. On \aigvd, at 1 and 2 gray levels, the attack still pushes clips slightly deeper into the range of real scores for Madry's objective than for \method (Wilcoxon $p=.012$ and $.006$). On \gvb, every model keeps almost all attacked clips up to 4 gray levels. Without weight averaging, the attack pushes \method's clips deeper at 1 and 2 gray levels on \aigvd (Wilcoxon $p=1\times10^{-4}$ at 1).

\begin{table}[t]
\centering
\caption{\textbf{Five folds.} Means over folds on \aigvd (AIGVD) and \gvb (GVB). Columns 1, 2, 4, and 8 give the TPR under evasion at that budget in gray levels. W/o averaging, \method without weight averaging.}
\label{tab:folds}
\small
\setlength{\tabcolsep}{2.5pt}
\begin{tabular}{l l cc cccc}
\toprule
& Model & AUC & TPR & 1 & 2 & 4 & 8\\
\midrule
\multirow{3}{*}{\rotatebox{90}{AIGVD}} & \method & .988 & .935 & .935 & .931 & .883 & .473\\
& w/o averaging & .984 & .917 & .903 & .889 & .788 & .440\\
& Madry & .987 & .933 & .924 & .919 & .867 & .479\\
\midrule
\multirow{3}{*}{\rotatebox{90}{GVB}} & \method & .999 & .998 & .998 & .993 & .993 & .888\\
& w/o averaging & .999 & .997 & .997 & .997 & .987 & .888\\
& Madry & .999 & .997 & .997 & .997 & .997 & .893\\
\bottomrule
\end{tabular}
\end{table}

\begin{table}[t]
\centering
\caption{\textbf{Detectors of an audit toolkit}~\cite{cakiroglu2026auditing}, with the columns of \cref{tab:main}. On CivicBench, columns give the clean AUC and TPR, the TPR under evasion, and the attacked false-alarm rate under framing at 1, 2, and 4 gray levels.}
\label{tab:audit}
\footnotesize
\setlength{\tabcolsep}{2.1pt}
\begin{tabular}{l cc ccc ccc c cccc c cccc}
\toprule
& \multicolumn{8}{c}{CivicBench (political)} && \multicolumn{4}{c}{\aigvd} && \multicolumn{4}{c}{\gvb}\\
\cmidrule{2-9}\cmidrule{11-14}\cmidrule{16-19}
& \multicolumn{2}{c}{clean} & \multicolumn{3}{c}{evasion $\uparrow$} & \multicolumn{3}{c}{framing $\downarrow$} &&
\multicolumn{2}{c}{clean} & \multicolumn{2}{c}{attacked} && \multicolumn{2}{c}{clean} & \multicolumn{2}{c}{attacked}\\
\cmidrule(lr){2-3}\cmidrule(lr){4-6}\cmidrule(lr){7-9}\cmidrule(lr){11-12}\cmidrule(lr){13-14}\cmidrule(lr){16-17}\cmidrule(lr){18-19}
Detector & AUC & TPR & 1 & 2 & 4 & 1 & 2 & 4 && AUC & TPR & 2 & 4 && AUC & TPR & 2 & 4\\
\midrule
\clipdet & .785 & .419 & .000 & .000 & .000 & 1.000 & 1.000 & 1.000 && .965 & .837 & .000 & .000 && .992 & .970 & .000 & .000\\
\flowts & .470 & .018 & .000 & .000 & .000 & 1.000 & 1.000 & 1.000 && .541 & .059 & .000 & .000 && .759 & .325 & .000 & .000\\
\codecmv & .415 & .028 & .001 & .000 & .000 & .696 & .725 & .739 && .634 & .101 & .000 & .000 && .820 & .366 & .006 & .004\\
\bottomrule
\end{tabular}
\end{table}

\section{Further Robustness Results}
\label{supp:attacks}

\paragraph{Stronger attacks} Stronger attacks find few evasions that PGD misses. Training with a single attack step can mask gradients~\cite{andriushchenko2020understanding}, so these checks also test for that. They test whether PGD, the main attack on \method, underestimates what an attacker can do. On CivicBench, APGD and Square attack \method and Madry's objective on the first 20 clips of each fixed attack set (\cref{supp:setup}), at 1, 2, and 4 gray levels. They evade or frame no clip that PGD does not. APGD also attacks every clip of these sets except Mochi's at 4 gray levels, and again at lower budgets where it succeeds. It breaks 1 of \method's 196 attacked generated clips beyond PGD, at 2 gray levels, and 1 of the 179 of Madry's objective. On \method's attacked clips of fold 0 of the general benchmarks, APGD finds an evasion at a lower budget than PGD on 4 of 79 clips. Square does so on 0 of 79. All of these evasions are at 8 gray levels. PGD with 1,000 steps instead of 100 finds no new evasion at 4 gray levels. At 8 gray levels, it finds 2 on 39 clips. TPRs under evasion at 8 gray levels are therefore upper bounds, while the stronger attacks change no TPR under evasion at 4 gray levels or below. For the other detectors, the fold-0 rates in \cref{tab:main} use the strongest attack on each clip. The five-fold rates use PGD alone for \method and for the baselines attacked on its clips, so they can be slightly too high. We use large budgets to test whether the gradient is masked. At 16 and 32 gray levels, PGD evades 97\% and 100\% of \method's attacked \aigvd clips. On \gvb, it evades 82\% and 98\%. This indicates that the gradient is informative and that the robustness at small budgets does not come from a masked gradient.

\paragraph{Uploads} Platforms re-encode uploaded video, which can undo an attack. To test this, every saved successful perturbation is added to the clean preprocessed clip. The clean and the attacked clip then pass through H.264 at CRF 23 and 28, where CRF is x264's constant rate factor and 23 its default. They also pass through a platform transcode that scales the clip to 360 pixels high, encodes it with H.264 at CRF 28, and scales it back. An evasion survives when the transformed attacked clip still scores below $\tau$, and a framing when it still scores at or above $\tau$. The transformed clean clip is a control that shows how often the transform alone changes the label.

\paragraph{Survival of the attacks} Re-encoding removes some attacks, and the platform transcode removes more. Against \stream, 74\% of the evasions and 53\% of the framings survive the default re-encode at CRF 23. On the clean clips, the re-encode alone changes the label of 19\% of these generated clips and 1\% of these real clips. After the platform transcode, 45\% of the evasions and 20\% of the framings survive. The framings of \restrav, \streammv, and \mvmob survive the default re-encode 12\% to 39\% of the time. Those of \method and of Madry's objective survive it every time (100\% and 100\%). The evasions that break \method also survive the default re-encode (84\%). They rarely survive the platform transcode (14\%, and 4\% for Madry's objective).

\paragraph{Attacks through the transcode} An attack through the platform transcode still frames real clips against \stream but adds no evasion or framing against \method. An attacker who expects the transcode can attack through it. We run Square with 1,000 queries per budget and score each query after the platform transcode, in the manner of expectation over transformation~\cite{athalye2018obfuscated}. Since our transcode is deterministic, this optimizes the transcoded clip directly. The attack runs on the first 40 clips of each fixed attack set at 1, 2, and 4 gray levels, against \method, Madry's objective, and \stream. It frames 38\% of real clips against \stream at 4 gray levels. Against \method it evades none of 34 generated clips and frames no clip beyond the 5\% that are flagged after the transcode alone. Against Madry's objective it frames 16\% of real clips at 4 gray levels, compared with 13\% that are flagged after the transcode alone.

\begin{table}[t]
\centering
\caption{\textbf{Paired tests over five folds} on \aigvd (AIGVD) and \gvb (GVB), per budget in gray levels. Lost, the clips that only one of the two models loses, as first:second. Wilcoxon $p$, signed-rank test on the catch rate (\cref{supp:ci}). Ours, \method. W/o avg., \method without weight averaging.}
\label{tab:paired}
\small
\setlength{\tabcolsep}{2.5pt}
\begin{tabular}{l l cccc}
\toprule
& Comparison & 1 & 2 & 4 & 8\\
\midrule
\multirow{4}{*}{\rotatebox{90}{AIGVD}} & ours/Madry, lost & 0:2 & 1:2 & 4:6 & 17:15\\
& \quad Wilcoxon $p$ & $.012$ & $.006$ & $.085$ & $.69$\\
& w/o avg./ours, lost & 2:0 & 5:1 & 20:6 & 15:12\\
& \quad Wilcoxon $p$ & $1\times10^{-4}$ & $2\times10^{-5}$ & $5\times10^{-4}$ & $.20$\\
\midrule
\multirow{2}{*}{\rotatebox{90}{GVB}} & ours/Madry, lost & 0:0 & 1:0 & 1:0 & 5:4\\
& \quad Wilcoxon $p$ & $.75$ & $.77$ & $.57$ & $.22$\\
\bottomrule
\end{tabular}
\end{table}

\paragraph{Perceptual cost} On CivicBench, the successful attacks change the clips little. We measure their change with PSNR (peak signal-to-noise ratio, higher is closer) and LPIPS (learned perceptual image patch similarity~\cite{zhang2018unreasonable}, lower is closer). At 1 gray level, their median PSNR is 51.1\,dB for evasion and 52.1\,dB for framing. Their median LPIPS is .001 and .002. At 4 gray levels, the median PSNR is 37.2\,dB for evasion and 37.4\,dB for framing. These medians pool 219 successful evasions and 222 successful framings. The median LPIPS is .025 and .025.

\paragraph{Framing on the general benchmarks} As on CivicBench, 1 gray level suffices to frame real clips against \stream, while framing against \method needs larger budgets. The attacks target real test clips of fold 0 that neither \method nor \stream flags, 40 on \aigvd and 40 on \gvb. PGD and APGD push 0\% and 5\% of the \aigvd clips over \method's threshold for 5\% false alarms at 1 and 2 gray levels. At 4 and 8 gray levels, they push 10\% and 45\%. On \gvb, they push 0\% to 45\% of the clips at these budgets. Against \stream, the three attacks PGD, APGD, and Square push all 20 of the \aigvd clips that they attack over the stricter threshold for 1\% false alarms at 1 gray level.

\paragraph{Attacks on a few frames} Attacks on one or three frames rarely evade \method, even at 32 gray levels. \method averages the embeddings of the 11 frames, so an attacker may instead spend a larger budget on a few frames. We attack one or three frames with PGD on \method's attacked clips of fold 0 of \aigvd, choosing the frames with the largest gradient. On three frames, attacks of 8 and 16 gray levels evade 3\% and 5\% of 39 clips. At 32 gray levels, they evade 8\%. On one frame, they evade 0\% at 8 gray levels and 0\% at 16. At 32, they evade 3\%. Averaging the frame embeddings likely explains these low rates, since it dilutes a change confined to a few frames (\cref{sec:method}).

\paragraph{Seeds} To measure the spread between training runs, we train \method three times on fold 0 of \aigvd. As mean $\pm$ standard deviation, the runs reach a clean AUC of .991 $\pm$ .001. Their clean TPR is .957 $\pm$ .004. Attacked on the same clips, their TPR under evasion is .941 $\pm$ .015 at 4 gray levels. At 8 gray levels, it is .540 $\pm$ .018.

\paragraph{Seeds on CivicBench} \method's gain under evasion holds across training runs. We also train each of three recipes three times on fold 0 of \aigvd. The recipes are \method, \method without its margin term, and Madry's objective with weight averaging. Each run is scored on CivicBench and attacked with PGD at 4 gray levels on the 117 paired fakes and 109 paired real clips that every run covers (\cref{tab:civic_seeds}). Under evasion, every run of \method keeps more attacked fakes flagged than every run of the other two recipes. The weakest run of \method keeps .684 of the attacked fakes flagged. The strongest run without the margin term keeps .675, and the strongest run of Madry's objective .675. Each run of \method also flags fewer attacked real clips than each run without the margin term, at most .248 against at least .257. Madry's objective flags about as many attacked real clips as \method, .235 against .235 on average.

\begin{table}[t]
\centering
\caption{\textbf{Seeds on CivicBench.} Three training runs per recipe on fold 0 of \aigvd, mean $\pm$ standard deviation. Clean AUC and TPR on all CivicBench clips. Under PGD at 4 gray levels, evasion is the share of attacked fakes that stay flagged and framing the share of real clips framed, on the paired clips that every run covers. Evasion is thus not the TPR under evasion of \cref{tab:ablation}.}
\label{tab:civic_seeds}
\footnotesize
\setlength{\tabcolsep}{1.6pt}
\begin{tabular}{l cccc}
\toprule
& \multicolumn{2}{c}{clean} & evasion $\uparrow$ & framing $\downarrow$\\
\cmidrule(lr){2-3}
Recipe & AUC & TPR & 4 & 4\\
\midrule
\method & .942 $\pm$ .007 & .784 $\pm$ .014 & .741 $\pm$ .041 & .235 $\pm$ .017\\
w/o margin term & .932 $\pm$ .004 & .753 $\pm$ .006 & .670 $\pm$ .004 & .266 $\pm$ .013\\
Madry & .927 $\pm$ .006 & .735 $\pm$ .009 & .650 $\pm$ .018 & .235 $\pm$ .011\\
\bottomrule
\end{tabular}
\end{table}

\paragraph{Detectors of an audit toolkit} Three detectors of a recent audit toolkit~\cite{cakiroglu2026auditing} fare like the published detectors of \cref{tab:main}. \Cref{tab:audit} gives their results. The toolkit is described only in a preprint. \clipdet is the toolkit's own linear probe on CLIP features. \flowts reads statistics of optical flow, and \codecmv reads statistics of codec motion vectors. On the general benchmarks, the table gives their clean AUC and TPR as means over five folds and their TPR under evasion at 2 and 4 gray levels. On CivicBench their AUC is .415 to .785. Evasion at 1 gray level works against all three. After framing at that budget, \clipdet and \flowts flag every attacked real clip.

\section{External Political Footage and Unseen Test Sets}
\label{supp:genvideo}

\paragraph{DF26} DF26~\cite{shykula2026df26} tests the detectors on real political footage from outside CivicBench. \Cref{tab:df26} gives the results of every detector on its official-statement scenario. The scenario pairs 71 real official statements with 213 text-to-video fakes of HunyuanVideo-1.5, LTX~2.3, and Wan2.2. The table also gives the AUC against all its fakes, which include image-to-video fakes that animate the first frame of a real clip. Each threshold gives 5\% false alarms on all real clips of DF26, not only on the official statements. On the statements alone, the share flagged before any attack can therefore differ from 5\%. For every detector it is higher (column 0). For framing, we attack every real statement with PGD at 1, 2, and 4 gray levels. \Cref{sec:depth} discusses the results.

\begin{table}[t]
\centering
\caption{\textbf{DF26, official statements.} AUC against the text-to-video fakes (T2V) and against all fakes (all), and the share of real statements flagged before the attack (0) and after framing at 1, 2, and 4 gray levels.}
\label{tab:df26}
\footnotesize
\setlength{\tabcolsep}{3pt}
\begin{tabular}{l cc cccc}
\toprule
& \multicolumn{2}{c}{AUC} & \multicolumn{4}{c}{framing $\downarrow$}\\
\cmidrule(lr){2-3}\cmidrule(lr){4-7}
Detector & T2V & all & 0 & 1 & 2 & 4\\
\midrule
CLIP probe & .455 & .459 & 15\% & 100\% & 100\% & 100\%\\
\restrav & .462 & .449 & 10\% & 100\% & 100\% & 100\%\\
D3 & .455 & .414 & 13\% & 100\% & 100\% & 100\%\\
\stream & .387 & .420 & 13\% & 100\% & 100\% & 100\%\\
\streammv & .404 & .424 & 10\% & 97\% & 99\% & 100\%\\
\mvmob & .161 & .278 & 17\% & 76\% & 83\% & 92\%\\
\midrule
Mavali, R2 & .427 & .457 & 18\% & 35\% & 52\% & 90\%\\
Mavali, R4 & .444 & .458 & 17\% & 35\% & 54\% & 83\%\\
Mavali, \dclip & .517 & .502 & 18\% & 23\% & 31\% & 37\%\\
Shahjahan et al. & .560 & .517 & 7\% & 99\% & 99\% & 99\%\\
Madry & .649 & .558 & 11\% & 18\% & 23\% & 24\%\\
\midrule
\method & .715 & .596 & 10\% & 14\% & 18\% & 25\%\\
\bottomrule
\end{tabular}
\end{table}

\paragraph{Other unseen test sets} Every detector loses accuracy away from its training data, and the detectors built to resist attacks transfer about as well as one another. We use three further test sets to measure how detectors transfer to unseen generators when they are trained on one benchmark and applied without changes. \Cref{tab:xbench} gives the detectors' AUC on \genbuster~\cite{wen2025busterx}, \vif~\cite{li2026skyra}, and \genvideo~\cite{chen2024demamba}. The Wild split of \genbuster includes recent closed generators such as Sora~2 and Veo~3.1. Every clip of the three test sets goes through the same re-encoding and sampling as the training data (\cref{supp:setup}). The clips of \genvideo's HotShot generator have only 8 frames, too few to sample 11, and are left out. The real clips of \genvideo are videos of MSR-VTT~\cite{xu2016msr}, a dataset of videos with captions. They have 3 frames per second, so their 11 frames span about 3.7 seconds.

\begin{table}[t]
\centering
\caption{\textbf{Unseen test sets.} AUC of detectors trained on fold 0 of \gvb (GVB) or \aigvd (AIGVD), or on both benchmarks (both). GenBuster splits are in-distribution (ID), out-of-distribution (OOD), and Wild. ViF is \vif, GenV.\ \genvideo. $^\dagger$Not built to resist attacks.}
\label{tab:xbench}
\small
\setlength{\tabcolsep}{3pt}
\begin{tabular}{c l ccccc}
\toprule
& & \multicolumn{3}{c}{GenBuster} & & \\
\cmidrule(lr){3-5}
& Detector & ID & OOD & Wild & ViF & GenV.\\
\midrule
\multirow{5}{*}{\rotatebox{90}{GVB}}
& \method & .873 & .961 & .808 & .703 & .962\\
& Madry & .871 & .948 & .810 & .704 & .968\\
& Mavali, R2 & .814 & .926 & .830 & .710 & .949\\
& CLIP probe$^\dagger$ & .899 & .932 & .861 & .687 & .964\\
& \stream$^\dagger$ & .775 & .751 & .738 & .766 & .927\\
\midrule
\multirow{5}{*}{\rotatebox{90}{AIGVD}}
& \method & .887 & .950 & .802 & .789 & .767\\
& Madry & .902 & .958 & .827 & .773 & .748\\
& Mavali, R2 & .814 & .924 & .759 & .740 & .701\\
& CLIP probe$^\dagger$ & .875 & .937 & .856 & .681 & .586\\
& \stream$^\dagger$ & .856 & .826 & .698 & .685 & .644\\
\midrule
both & \method & .886 & .956 & .799 & .789 & .939\\
\bottomrule
\end{tabular}
\end{table}

On unseen generators, every detector built to resist attacks is less robust than on its training benchmark (\cref{tab:xattack}). We train \method, Madry's objective, and the CLIP probe on fold 0 of \gvb and attack them on the same clips of the Wild split of \genbuster and of \vif. Each threshold gives 5\% false alarms on the real clips of its test set. On the Wild split, \method keeps a TPR under evasion of .304 at 2 gray levels and Madry's objective .351. On their shared attacked clips, no difference between the two is significant, on the Wild split or on \vif, at any budget (McNemar $p\geq.38$).

\Cref{prop:survival} links this weakening to the margin. A clip with a smaller margin breaks at a smaller budget. The threshold is set anew on the real clips of each test set. On the Wild split, the median margin of the flagged generated clips falls to 33\% of its in-domain value for \method and to 19\% for Madry's objective.

\begin{table}[t]
\centering
\caption{\textbf{More variants on CivicBench}, with the columns, clips, and marks of \cref{tab:ablation}. Bold marks the best value over both tables.}
\label{tab:ablation_civic_more}
\footnotesize
\setlength{\tabcolsep}{2.4pt}
\begin{tabular}{l cc cc cc}
\toprule
& \multicolumn{2}{c}{clean} & \multicolumn{2}{c}{evasion $\uparrow$} & \multicolumn{2}{c}{framing $\downarrow$}\\
\cmidrule(lr){2-3}\cmidrule(lr){4-5}\cmidrule(lr){6-7}
Variant & AUC & TPR & 2 & 4 & 2 & 4\\
\midrule
\method & .941 & .779 & .692 & .593 & .138 & .211\\
TRADES + margin & .923 & .739 & .678$^{\dagger}$ & .564$^{\dagger}$ & .165$^{\dagger}$ & .294\\
MART + margin & .937 & .756 & .688$^{\dagger}$ & .580$^{\dagger}$ & .156$^{\dagger}$ & .239$^{\dagger}$\\
w/o preservation & .935 & .759 & .703$^{\dagger}$ & .572$^{\dagger}$ & .128$^{\dagger}$ & .266$^{\dagger}$\\
w/o generator weights & .921 & .741 & .661$^{\dagger}$ & .536 & .128$^{\dagger}$ & .239$^{\dagger}$\\
\bottomrule
\end{tabular}
\end{table}

\section{Further Ablations}
\label{supp:ablation_aigvd}

\begin{table}[t]
\centering
\caption{\textbf{Robustness on unseen test sets.} TPR on \genbuster's Wild split and on \vif (ViF), clean and under evasion at 1, 2, and 4 gray levels. Models are trained on fold 0 of \gvb or on both benchmarks (both). $^\dagger$Not built to resist attacks.}
\label{tab:xattack}
\small
\setlength{\tabcolsep}{3pt}
\begin{tabular}{l l cccc}
\toprule
Set & Detector & Clean & 1 & 2 & 4\\
\midrule
\multirow{4}{*}{Wild} & \method & .380 & .352 & .304 & .228\\
& Madry & .373 & .362 & .351 & .283\\
& CLIP probe$^\dagger$ & .420 & .000 & .000 & .000\\
& \method, both & .313 & .298 & .282 & .235\\
\midrule
\multirow{4}{*}{ViF} & \method & .287 & .237 & .194 & .158\\
& Madry & .252 & .225 & .207 & .180\\
& CLIP probe$^\dagger$ & .206 & .000 & .000 & .000\\
& \method, both & .462 & .416 & .370 & .324\\
\bottomrule
\end{tabular}
\end{table}

\paragraph{On CivicBench} Added to TRADES or MART, the margin term raises the TPR under evasion. \Cref{tab:ablation_civic_more} adds four variants to \cref{tab:ablation}, with the same columns and clips. Two add the margin term to TRADES and to MART, to test whether it also helps other objectives. The other two remove a part of \method, its preservation term or its generator weights. For MART, the TPR under evasion rises from .598 to .688 at 2 gray levels. At 4 gray levels, it rises from .492 to .580. For TRADES, it rises from .634 to .678 at 2 gray levels and from .540 to .564 at 4. MART with the margin term is not significantly worse than \method in any column. TRADES with it still flags more attacked real clips than \method at 4 gray levels ($p<.001$). Without the preservation term, slightly more real clips are framed at 4 gray levels, but not significantly ($p=.070$). The rate is .266 against .211 for \method. Without the generator weights, the TPR under evasion at 4 gray levels drops to .536 ($p=.008$). None of the four differs significantly from \method at 2 gray levels. Two training choices of \cref{tab:ablation} matter at the larger budget. Attacking only generated clips in training gives the highest clean TPR. At 4 gray levels, however, this variant's TPR under evasion is lower ($p=.005$). Without weight averaging, more real clips are framed at 4 gray levels ($p=.037$).

\paragraph{A margin on real clips} Adding the margin term on real clips fails a rule that we fixed before training. \method's margin term acts on generated clips only. The same term on attacked real clips might lower framing too. We trained this two-sided variant once, with \method's recipe and data. Before training, we fixed a rule for keeping it. The variant is kept only if it flags fewer attacked real clips at 2 gray levels, is not significantly worse under evasion at 1, 2, or 4 gray levels, and loses at most .005 of clean AUC. At 2 gray levels it flags fewer attacked real clips, 11.4\% against 14.1\%, but not significantly ($p=.074$). At 4 gray levels, it flags 24.3\% against 22.2\% ($p=.345$). Its clean AUC is .943 against .943. Its clean TPR, however, is lower, .757 against .787. Under evasion at 4 gray levels, 29.3\% of its attacked fakes evade. For \method, 20.1\% evade ($p<.001$). The variant thus fails the rule, so we keep the one-sided margin of \method. In short, the margin on generated clips helps against evasion, while the margin on real clips did not measurably lower framing. The one-sided term acts only on attacked fakes, so its effect on framing in \cref{tab:ablation} is indirect.

\paragraph{Fine-tuning without attacks} The training attack adds robustness beyond the robust encoder, most clearly at 4 gray levels. \method fine-tunes the robust encoder \dclip on attacked clips. Its robustness could instead come from the fine-tuning alone. To separate the two, we trained a control once with \method's recipe and data, but without the training attack and without the margin term. Before training, we fixed four predictions. On the same attacked fakes, the control should keep at least 10 percentage points fewer flagged than \method, both at 2 and at 4 gray levels. On the same attacked real clips, it should flag at least 10 points more at 2 gray levels. Its clean AUC should lie within .03 of \method's. The clean AUC holds, at .914 against .941. On the 104 attacked fakes that both flag, 33.3\% of the control's fakes evade at 4 gray levels. For \method, 15.1\% evade ($p<.001$). At 2 gray levels, the gap is 8.7 points ($p=.004$). This gap is significant but below the 10 points we predicted. At 2 gray levels, the robust encoder thus carries most of the robustness. On the 109 real clips framed at 2 gray levels, the control flags 20.2\% and \method flags 13.8\% ($p=.002$). This gap of 6.4 points is also significant but below the 10 points we predicted.

\paragraph{Encoder and recipe} Both the encoder and the recipe matter, and the recipe carries over to a smaller encoder. \method's robustness could come from its encoder or from its training recipe. To separate the two, we run both recipes on the same encoder (\cref{tab:encswap}). The recipe of Mavali et al.~\cite{mavali2026adversarial} trains a linear probe on the embeddings of a robust encoder such as R4 (\cref{supp:setup}). All models are attacked with PGD on \method's attacked clips. On R4, \method's recipe keeps a higher TPR under evasion than that of Mavali et al., on the 162 clips of \aigvd and 185 of \gvb that both models flag. On \aigvd, the gain is +.236 and +.282 at 1 and 2 gray levels. At 4 and 8 gray levels, it is +.367 and +.129. On \gvb, it is +.081 to +.334. Its clean TPR on \aigvd is also higher, by +.129. Moving \method's recipe from R4 to \dclip, a larger encoder, adds +.045 and +.108 on \aigvd at 1 and 2 gray levels. At 4 and 8 gray levels, it adds +.254 and +.370. These gains again use the clips that both models flag. Every difference is paired over five folds with bootstrap $p\leq.001$.

\begin{table}[t]
\centering
\caption{\textbf{Encoder and recipe}, mean over five folds. Clean AUC and TPR, and TPR under evasion at 1 to 8 gray levels. R4 is the ViT-L/14 encoder of Schlarmann et al.~\cite{schlarmann2024robust}, adversarially fine-tuned at 4 gray levels. \dclip is \method's ViT-H/14 encoder.}
\label{tab:encswap}
\small
\setlength{\tabcolsep}{2.3pt}
\begin{tabular}{l l cc cccc}
\toprule
& Recipe on encoder & AUC & TPR & 1 & 2 & 4 & 8\\
\midrule
\multirow{3}{*}{\rotatebox{90}{AIGVD}}
& Mavali, R4 & .952 & .785 & .675 & .584 & .335 & .019\\
& \method on R4 & .983 & .914 & .886 & .823 & .639 & .128\\
& \method on \dclip & .988 & .935 & .935 & .931 & .883 & .473\\
\midrule
\multirow{3}{*}{\rotatebox{90}{GVB}}
& Mavali, R4 & .988 & .949 & .899 & .852 & .663 & .152\\
& \method on R4 & .997 & .990 & .980 & .965 & .905 & .460\\
& \method on \dclip & .999 & .998 & .998 & .993 & .993 & .888\\
\bottomrule
\end{tabular}
\end{table}

\paragraph{On \aigvd} Fine-tuning again matters most. \Cref{tab:ablation_aigvd} changes one part of \method at a time on the general benchmark \aigvd, with thresholds at 5\% false alarms. Without fine-tuning, the detector is the linear probe of Mavali et al.\ on \dclip. It detects .727 of the clean generated clips against .957. At 8 gray levels, it keeps .226 against .540. On fold 0, where all rows are attacked on the same 40 clips, each of the other changes alters the attack outcome of at most 3 of them. The standard deviation over three training runs of \method is below one clip (\cref{supp:attacks}). Three runs say little about the spread, however, so we judge these parts by the clean TPR and by five-fold runs. Weight averaging raises the clean TPR and keeps more attacked clips flagged at 1 and 2 gray levels over five folds (\cref{supp:folds}).

With weight averaging, a standard adversarial loss comes close to \method when it is used to fine-tune the same encoder (bottom of \cref{tab:ablation_aigvd}). To test the margin term, we add it to three objectives with weight averaging, namely \method without its margin term, TRADES, and MART. It raises the TPR under evasion of \method without its margin term by +.028 at 2 gray levels ($p=.002$). It raises that of TRADES by +.053 at 8 gray levels ($p=.042$). It makes no significant difference for MART.

\begin{table}[t]
\centering
\caption{\textbf{Ablations on \aigvd.} TPR, clean and under evasion with PGD at 1 to 8 gray levels. Top, \method with one part changed, on fold 0. Bottom, means over five folds with paired tests. $^{\dagger}$Not significantly different from \method. $^{*}$Differs from the row above at $p<.05$.}
\label{tab:ablation_aigvd}
\small
\setlength{\tabcolsep}{2.3pt}
\begin{tabular}{l ccccc}
\toprule
Variant & Clean & 1 & 2 & 4 & 8\\
\midrule
\method & .957 & .957 & .957 & .932 & .540\\
without weight averaging & .926 & .903 & .879 & .810 & .486\\
attacking generated clips only & .959 & .959 & .959 & .910 & .492\\
without preservation term & .968 & .968 & .944 & .895 & .532\\
without fine-tuning (probe) & .727 & .702 & .602 & .577 & .226\\
\midrule
\method w/o margin & .931 & .931 & .903 & .860 & .478\\
\quad + margin (\method) & .935 & .935 & .931$^{*}$ & .883 & .473\\
Madry & .933$^{\dagger}$ & .924$^{\dagger}$ & .919$^{\dagger}$ & .867$^{\dagger}$ & .479$^{\dagger}$\\
TRADES & .935$^{\dagger}$ & .921$^{\dagger}$ & .906 & .859$^{\dagger}$ & .457$^{\dagger}$\\
\quad + margin & .939 & .925 & .915 & .873 & .508$^{*}$\\
MART & .939$^{\dagger}$ & .939$^{\dagger}$ & .930$^{\dagger}$ & .882$^{\dagger}$ & .447$^{\dagger}$\\
\quad + margin & .938 & .933 & .929 & .858 & .476\\
\bottomrule
\end{tabular}
\end{table}

\section{Compute and Broader Impact}
\label{supp:compute}

\paragraph{Compute} Generating CivicBench took about 730 GPU hours. Scoring it with every detector took about 450, and attacking it about 4,300. We used A100 GPUs with 40\,GB of memory and V100 GPUs with 32\,GB. Training \method on one fold takes about 0.5 hours on one A100 or 1.2 hours on a V100. Scoring a clip of 11 frames takes about 0.8 seconds, decoding included. The whole project used about 19,600 GPU hours, including the general benchmarks and runs that failed or were cancelled. Most of them went to attacks, since each attacked clip costs minutes of gradient steps or queries at every budget.

\paragraph{Broader impact} We built CivicBench and \method to make detectors of political video harder to turn against the public record in either direction, by hiding generated clips or by framing real footage. The same results also tell attackers which budgets still work. We report them so that those who deploy a detector can choose its false-alarm rate and the budget that it must withstand. \Cref{supp:civic} describes the release and the ethics of the footage.

\end{document}